\documentclass{article}

\usepackage{microtype}
\usepackage{graphicx}
\usepackage{subcaption}
\usepackage{booktabs} %

\usepackage{hyperref}

\usepackage[preprint]{icml2026}

\usepackage{amsmath}
\usepackage{amssymb}
\usepackage{mathtools}
\usepackage{amsthm}
\usepackage{thmtools}
\usepackage{paralist}

\usepackage[capitalize,noabbrev]{cleveref}

\crefname{section}{Section}{Sections}
\Crefname{section}{Section}{Sections}

\crefname{appendix}{Appendix}{Appendices}
\Crefname{appendix}{Appendix}{Appendices}

\theoremstyle{plain}
\newtheorem{theorem}{Theorem}[section]
\newtheorem{proposition}[theorem]{Proposition}

\theoremstyle{definition}

\theoremstyle{remark}

\usepackage[textsize=tiny]{todonotes}

\usepackage{thm-restate}
\usepackage{booktabs}
\usepackage{wrapfig}
\usepackage{multirow}
\usepackage{enumitem}
\usepackage{caption}
\usepackage[dvipsnames,table]{xcolor}
\usepackage{listings}

\def\1{\bm{1}}

\DeclareMathAlphabet{\mathsfit}{\encodingdefault}{\sfdefault}{m}{sl}
\SetMathAlphabet{\mathsfit}{bold}{\encodingdefault}{\sfdefault}{bx}{n}

\icmltitlerunning{Uncertainty Quantification for LLMs Fails under Ambiguity}

\begin{document}

\twocolumn[
  \icmltitle{The Illusion of Certainty: Uncertainty Quantification for LLMs Fails under Ambiguity}

  \icmlsetsymbol{equal}{*}

  \begin{icmlauthorlist}
    \icmlauthor{Tim  Tomov}{tum}
    \icmlauthor{Dominik Fuchsgruber}{tum}
    \icmlauthor{Tom Wollschläger}{tum}
    \icmlauthor{Stephan Günnemann}{tum}
  \end{icmlauthorlist}

  \icmlaffiliation{tum}{School of Computation, Information and Technology \& Munich Data Science Institute - Technical University of Munich}

  \icmlcorrespondingauthor{Tim Tomov}{tim.tomov@tum.de}

  \icmlkeywords{Machine Learning, ICML}

  \vskip 0.3in
]

\printAffiliationsAndNotice{}  %

\begin{abstract}
    Accurate uncertainty quantification (UQ) in Large Language Models (LLMs) is critical for trustworthy deployment.
    Despite ubiquitous ambiguities in real-world language -- so-called aleatoric uncertainty -- existing UQ methods are typically benchmarked against tasks with unambiguous answers. In this work, we
    formally establish the role of aleatoric uncertainty in existing uncertainty estimation paradigms. We theoretically explain why they perform well under zero aleatoric uncertainty and why their performance degrades on ambiguous data. To empirically demonstrate our results, we introduce MAQA\(^*\) and AmbigQA\(^*\), the first ambiguous question-answering (QA) datasets equipped with ground-truth answer distributions estimated from factual co-occurrence, and find that methods degrade to near-random performance.
    Our work motivates a paradigm shift to explicitly account for ambiguities in the responses when developing UQ methods for LLMs.
    
\end{abstract}

\section{Introduction}

Many linguistic tasks that are solved by Large language models (LLMs) can be framed as \emph{question-answering} (QA): a user poses a query, and the model provides an answer. As LLMs are increasingly deployed in high-stakes domains such as medical diagnosis, legal advice, and autonomous decision-making, it becomes critical to understand when a model’s prediction should be trusted. This is closely tied to having meaningful estimates of how well the model understands the data, commonly referred to as \emph{epistemic uncertainty}. Importantly, when assessing model reliability, questions can have multiple correct answers. This creates ambiguity about the expected correct answer. Consider these two examples:

\begin{itemize}[label={}, leftmargin=*]
    \item \textbf{Single-answer (No ambiguity):} ``\emph{Which hormone do I lack if I have type 1 diabetes?}'' $\rightarrow$ \textit{Insulin}.
    \item \textbf{Multi-answer (Ambiguity):} ``\emph{Which   medication should I take for type 2 diabetes?}'' $\rightarrow$ \textit{Metformin, Sulfonylureas, DPP-4 Inhibitors, ...} (all plausible, but differently likely).
\end{itemize}

The first example has only one correct answer, and an LLM's predictive distribution over possible responses should put all mass on this answer. In the second example, multiple answers are correct, each with a different likelihood. This is known as \emph{aleatoric uncertainty}: It refers to the randomness that is intrinsic to the distribution of true answers itself. Importantly, we focus on settings in which ambiguity is \emph{not resolvable}. In contrast, ambiguity arising from under-specified questions can be mitigated by clarifications \citep{hou2024decomposinguncertaintylargelanguage,walha2025finegraineduncertaintydecompositionlarge,kirchhof2025positionuncertaintyquantificationneeds}, while the ambiguity considered in this work is inherent to the task itself and, by definition, is irreducible. Such ambiguity is not a rare edge case: many common language-generation tasks inherently admit multiple valid outputs, including question answering with multiple correct answers \cite{yang2025maqaevaluatinguncertaintyquantification}, summarization \cite{koupaee2025faithfulunfaithfulambiguousmultiagent}, machine translation \cite{wu2025multiplereferencesmeaningfulvariations}, and code generation \cite{maveli2025largelanguagemodelscapture}.

\begin{figure*}[t!]
  \centering
  \includegraphics[width=\linewidth]{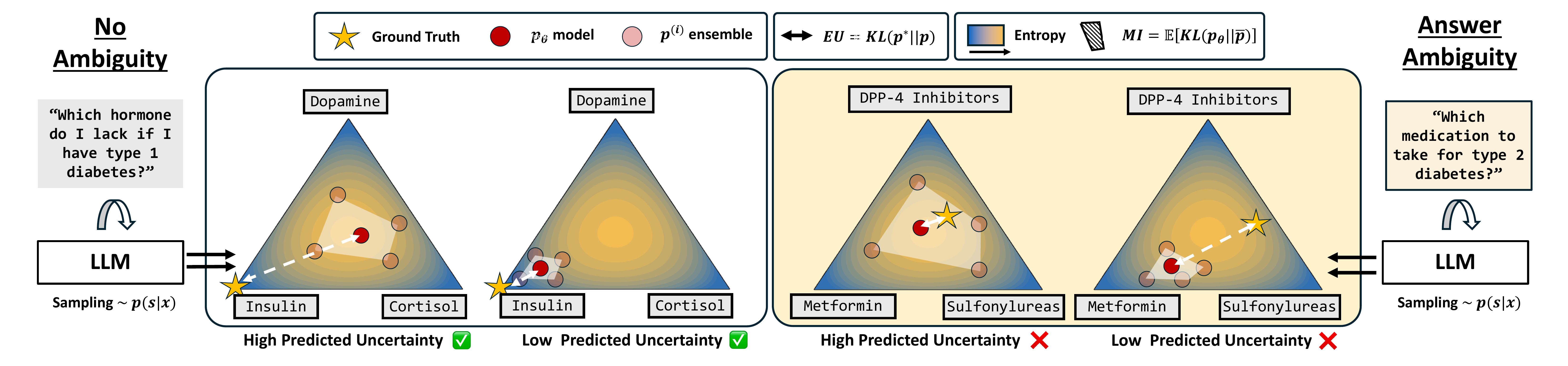}
    \caption{Epistemic uncertainty estimation with and without answer ambiguity. \textbf{Left (No ambiguity):} When the ground-truth distribution $p^*$ is concentrated on a single answer ($H(p^*)=0$), predictive entropy and mutual information reliably indicate epistemic uncertainty. \textbf{Right (Answer ambiguity):} When multiple answers are valid, $p^*$ is unconstrained. In this regime, both entropy and mutual information lose a consistent interpretation: high and low values may correspond to both low and high epistemic uncertainty.}
  \label{fig:simplex}
    \vspace{-1em}
\end{figure*}

Despite this, most existing uncertainty quantification (UQ) methods for LLMs are evaluated almost exclusively in settings resembling the first question, where aleatoric uncertainty is zero \citep{devic2025calibrationcollaborationllmuncertainty}. Under these strong assumptions, many methods demonstrate strong performance in estimating epistemic uncertainty \citep{kuhn2023semanticuncertaintylinguisticinvariances,duan2024shifting,yadkori2024believebelievellm}. 
Notably, their success is empirically demonstrated and, so far, lacks a principled theoretical explanation. This raises the question of whether the observed effectiveness of current UQ methods is a consequence of fundamental algorithmic properties, or merely a byproduct of the absence of aleatoric uncertainty in which they are typically evaluated. We systematically address this by asking: \emph{What role does aleatoric uncertainty play for uncertainty in LLMs both theoretically and empirically?}

To answer this, we introduce MAQA\(^*\) and AmbigQA\(^*\), the first ambiguous QA datasets equipped with explicit ground-truth answer distributions \(p^*\), estimated from factual co-occurrence statistics (\Cref{sec:dataset}). These datasets enable, for the first time, a principled evaluation of uncertainty estimators under measurable real-world ambiguity.
We contrast the absence and presence of aleatoric uncertainty theoretically and empirically. To that end, we organize UQ approaches into families based on the information they utilize: \emph{Consistency}, \emph{Internal Representation}, and \emph{Ensemble} based.
For the consistency and ensemble families, we formally show why they necessarily succeed at UQ under zero aleatoric uncertainty (\Cref{sec:zero_au}), consistent with empirical evidence from existing literature that (implicitly) considers such settings. Building on this, we then lay out why the \emph{presence of aleatoric uncertainty is detrimental for existing uncertainty estimators theoretically}. A systematic evaluation on our novel dataset confirms that their performance degrades to close-to-random under aleatoric uncertainty (\Cref{sec:non_trivial_au}).

\section{Background}
\label{sec:background}

\begin{table*}[t]
  \caption{Examples of question-answer-distribution pairs}
    \vspace{-0.2cm}
  \label{tab:dataset_examples}
  \centering
  \resizebox{\textwidth}{!}{%
    \begin{tabular}{lccccc}
      \toprule
      \textbf{Dataset}   & \textbf{Question}                                                 
                         & \textbf{Answer(s)}          & {\# \textbf{Counts} in Data}    
                         & \(\mathbf{p^*}\)           & \textbf{Entropy} \(\mathbf{H(p^*)}\) \\
      \midrule
      TriviaQA          & Where in England was Dame Judi Dench born?                       
                         & \(\{Yorkshire\}\)           &          \text{n/a
                         }          
                         & \([1.00]\)                  & 0.0               \\
                         \midrule
      MAQA\(^*\)            & What is one essential component of the fire triangle?  %
                         & \(\{Heat,Fuel,Oxygen\}\)               & \(\{31,32,25\}\)                  
                         & \([0.35,0.36,0.29]\)                        & 1.1           \\
      AmbigQA\(^*\)          & What is the name of one princess in Frozen?                      
                         & \(\{Elsa,Anna\}\)           & \(\{188,91\}\)      
                         & \([0.67,0.33]\)            & 0.63           \\
      \bottomrule
    \end{tabular}
  }
\vspace{-0.3cm}
\end{table*}

Uncertainty quantification (UQ) in machine learning characterizes the uncertainty in a model's predictive distribution for a given input \(x\). This uncertainty, often referred to as \emph{total uncertainty}, stems from two distinct sources: \emph{epistemic uncertainty}, reflecting uncertainty in the model itself due to limited training data, model misspecification, or optimization artifacts, and \emph{aleatoric uncertainty}, which represents the intrinsic randomness in the true data-generating process \citep{H_llermeier_2021, gawlikowski2022surveyuncertaintydeepneural}. Epistemic uncertainty can be reduced with sufficient data and a well-specified model, whereas aleatoric uncertainty is irreducible by definition. Isolating \emph{epistemic} uncertainty is central to decision making, as it enables determining when a prediction can be trusted or not. When both sources are present, they jointly shape the model’s predictive distribution. Naive uncertainty estimates may conflate epistemic uncertainty with genuine data ambiguity. Disentangling uncertainty is therefore a central challenge in reliable ML.

With the general capability of LLMs to address diverse tasks by framing them as question-answering (QA) problems \citep{sanh2022multitaskpromptedtrainingenables}, a natural approach to uncertainty quantification in LLMs is assessing the model's confidence in the answers it provides through their semantic variability. However, reasoning about uncertainty 
in natural-language requires an appropriate representation of responses, as syntactical variations do not necessarily correspond to meaningful semantic differences that reflect the model's belief. Therefore, it is useful to group generated answers into semantic equivalence classes \citep{kuhn2023semanticuncertaintylinguisticinvariances}. For instance, to the question ``What is the capital of France?'', the answers ``Paris'' or ``The capital is Paris'' correspond to the same semantic class. Like previous work, we study the distribution over these semantically distinct classes, denoted \(p\) in the remainder (implementation details are given in \Cref{app:implementation}). This distributional perspective is implicitly adopted by many consistency-based uncertainty estimation methods \citep{kuhn2023semanticuncertaintylinguisticinvariances,duan2024shifting,nikitin2024kernellanguageentropyfinegrained,yadkori2024believebelievellm}. Framing uncertainty quantification like this allows treating UQ for LLMs as a classification problem over semantic answers. For our subsequent analysis, we categorize existing UQ estimators based on the information they use:
\begin{inparaenum}[(i)]
    \item \textbf{Consistency}: methods that rely on some variation of the output, typically quantifying epistemic uncertainty via variation measures such as entropy.
    \item \textbf{Ensembles}: Bayesian-inspired methods that approximate a posterior in the model parameter space by aggregating predictions from multiple models.  
    \item \textbf{Internal Representations}: methods that probe the hidden states of the LLM.
\end{inparaenum}

Most existing UQ methods for LLMs do not explicitly target epistemic uncertainty. However, they are evaluated through downstream decision-making proxies, such as separating correct from incorrect answers. Such metrics implicitly assume a binary notion of correctness and thus effectively treat uncertainty as a surrogate for the model's epistemic error.
This implicitly assumed equivalence of predictive and epistemic uncertainty breaks down once aleatoric uncertainty enters the picture: when a question admits multiple valid answers, correctness is no longer binary but must be described at a distributional level. To capture this distinction, we follow \citet{kotelevskii2025riskuncertaintygeneratingpredictive} and define the \emph{total uncertainty (TU)} as cross-entropy between the true distribution  \(p^*\) over semantic classes and the semantic distribution predicted by the model \(p\). This allows following decomposition: \emph{Aleatoric uncertainty (AU)} is the entropy of the true distribution \(p^*\) and \emph{epistemic uncertainty (EU)} the KL divergence between \(p^*\) and the predicted distribution  \(p\)\footnote{We assume the model class is expressive enough to represent \(p^*\), so any mismatch between \(p\) and \(p^*\) is, in principle, reducible.}
\begin{equation}
\underbrace{\mathrm{CE}(p^*, p)}_{\text{Total (TU)}} 
= \underbrace{H(p^*)}_{\text{Aleatoric (AU)}} 
+ \underbrace{\mathrm{KL}(p^* \| p)}_{\text{Epistemic (EU)}}
\label{eq:decomposition}
\end{equation}
Consequently, epistemic uncertainty directly captures the discrepancy between the model’s beliefs and the true answer's distribution. This aligns with the decision-making objectives implicitly targeted by prior evaluations, while remaining well-defined under aleatoric uncertainty.

\section{Uncertainty Quantification And Aleatoric Uncertainty}
\label{sec:theory}

We first formally examine how the absence or presence of aleatoric uncertainty affects existing UQ paradigms. To that end, we focus on two prominent proxies: The entropy of the predictive distribution and the mutual information between the predicted target variable and model parameters in an ensemble of LLMs. Our results generalize to conceptually similar approaches, e.g., all consistency-based methods.

\subsection{Existing Uncertainty Quantification Provably Works Under No Aleatoric Uncertainty}
\label{sec:zero_au}

We first revisit the case that is (implicitly) studied by existing work in which the ground-truth has no aleatoric uncertainty. In particular, we formally approach the empirical strong performance of existing UQ approaches under this assumption.
If aleatoric uncertainty is zero, i.e. \(H(p^*)=0\), then the epistemic uncertainty reduces to the negative log-probability the model assigns to the correct semantic class: \(EU = -\log p(y=y^*)\)(\Cref{prop:zero_au}). Therefore, epistemic uncertainty can be directly understood as the model’s confidence in the correct answer. Geometrically, this means that the true distribution \(p^*\) must be located at one of the vertices of the probability simplex (\Cref{fig:simplex}, left). This constraint on \(p^*\) provides the basis for two complementary theoretical results that apply to consistency-based and ensemble estimators, explaining their empirical success as a necessary consequence of zero aleatoric uncertainty.

\subsubsection{Why Consistency-based Methods Work under Zero Aleatoric Uncertainty}
\label{subsec:theo_var_zero}
Methods based on output consistency quantify uncertainty through variation in the predictive distribution \(p\).
We show that high predictive entropy in \(p\) necessarily results in high epistemic uncertainty and low entropy implies low epistemic uncertainty with high probability. This gives a principled explanation of why the predictive entropy is useful for estimating epistemic uncertainty when aleatoric uncertainty is absent. As other variation-based quantities computed from \(p\) correlate well with its entropy, these results also translate to other consistency-based approaches.

\begin{restatable}[name=High Entropy $\Rightarrow$ High EU]{theorem}{thmhighentropy}
\label{thm:high_entropy}
Let there be \(K\!\ge\!2\) classes and \(\delta\in[0,\log K]\) be a threshold on the entropy indicating uncertainty. Furthermore, let \(\alpha_\delta\) be the maximal possible probability on some class for this entropy threshold \(H(p)\geq\delta\). Then the epistemic uncertainty with \(H(p)\ge\delta\) is at least:
\[
EU = \mathrm{KL}(p^* \| p) \;\ge\;-\log\alpha_\delta.
\]
\end{restatable}
High entropy \(H(p)\ge\delta\) implies that the predictive distribution must become increasingly less concentrated. Therefore, the maximum probability assigned to any class can be at most \(\alpha_{\delta}\). Naturally, this also holds for the correct ground-truth semantic class \(y^*\). Since epistemic uncertainty is quantified as \(-\log p(y=y^*)\), this flat predictive distribution hence leads to large epistemic uncertainty (\Cref{fig:simplex} left-most). Thus, \Cref{thm:high_entropy} shows that \emph{high predictive entropy necessarily implies high epistemic uncertainty}.
\begin{restatable}[name=Low Entropy $\Rightarrow$ Low EU with High Probability]{theorem}{thmlowentropy}
\label{thm:low_entropy}
Let there be \(K\!\ge\!2\) classes and \(\delta\in[0,\log 2]\) be a threshold on the entropy indicating uncertain outputs. Furthermore let
\( \bar{\mathcal{L}}\;=\;\mathbb{E}_{(x,y)}[-\log p_y] \) be the model's average loss and \(\gamma_{\delta}\) be the minimal maximal confidence in a prediction \(p\) for the given entropy threshold \(H(p)\leq\delta\). Then the probability that the epistemic uncertainty with \(H(p)\leq\delta\) will be less than \(-\log(\gamma_{\delta})\) satisfies:
\begin{equation*}
\begin{aligned}
\Pr\!\bigl(
EU &\leq -\log\gamma_{\delta}\mid H(p)\leq \delta
\bigr)
\\
&\geq 1 - \frac{\bar{\mathcal{L}}}
{-\log(1-\gamma_{\delta})\,\Pr(H(p)\leq\delta)} .
\end{aligned}
\end{equation*}
\end{restatable}
\Cref{thm:low_entropy} complements \Cref{thm:high_entropy} by showing that low entropy implies low epistemic uncertainty with high probability.
When the predictive entropy is small, i.e. \(H(p)\le\delta\), most of the probability mass must lie on a single class with weight at least \(\gamma_\delta\). This induces a dichotomy: if the class is correct, epistemic uncertainty is small (\(\le -\log \gamma_\delta\)); if incorrect, it is large (\(\ge -\log(1-\gamma_\delta)\)) (\Cref{fig:simplex} second-to-left). A deterministic upper bound is therefore impossible, but we can obtain a probabilistic guarantee that depends on the model's average performance. Noting that the training loss \(-\log p_y\) coincides with epistemic uncertainty under zero AU (\Cref{prop:zero_au}), the average loss \(\mathcal{L}\) coincides with the expected EU. For a well-trained model with small average \(\mathcal{L}\), frequent high-EU errors are hence unlikely to occur. In other words, highly confident but incorrect predictions must occur not too often. Since our bound depends on the probability with which the model makes confident predictions (in terms of entropy), the guarantee may become loose if
the model is predominantly not confident. In practice, however, models are trained toward confident predictions, and our bound becomes tight. Intuitively, \Cref{thm:low_entropy} shows that for models that are likely to make confident predictions and perform well on average, \emph{observing a low predictive entropy corresponds likely to a low epistemic uncertainty}.
The strict lower and probabilistic upper bound force a correlation between predictive semantic entropy and the true epistemic uncertainty. Therefore, entropy is a useful proxy for epistemic uncertainty \emph{if no aleatoric uncertainty is present}.

\subsubsection{Why Ensemble-based Methods Work Under Zero Aleatoric Uncertainty}  
\label{subsec:theo_ens_zero}

Showing that zero AU enables predictive entropy to be an effective estimate of the true epistemic uncertainty has direct implications for ensemble-based UQ. For ensembles, EU is estimated as the mutual information (MI) between the model parameters and the predicted target variable.
\begin{equation}
    \underbrace{\mathrm{MI}(\bar{p}; \theta)}_{\text{Estimated EU}} 
= H(\bar{p}) 
- \mathbb{E}_\theta\left[H(p_\theta)\right] \leq H(\bar{p}),
\label{eq:mutualinformation}
\end{equation}
where \(\bar{p} = \mathbb{E}_\theta\left[p_\theta\right]\) is the Bayesian model average that serves as the ensemble's prediction. \Cref{eq:mutualinformation} shows that the MI, which estimates EU, is bounded by the entropy of the ensemble's predictive distribution. Therefore, a large MI implies a large predictive entropy in \(\bar{p}\) which, in turn, leads to a high true epistemic uncertainty as per \Cref{thm:high_entropy} (\Cref{fig:simplex}, left-most). Therefore, mutual information also admits a formal guarantee regarding the true epistemic uncertainty: \emph{If AU is zero, large mutual information necessarily implies a high true epistemic uncertainty}.

Similar to consistency-based estimators, low mutual information does not guarantee a low epistemic uncertainty in a deterministic way. However, if the individual predictors sampled from \(p_\theta\) achieve low expected error, then these predictors assign high probability to the correct label, again resulting in near-zero entropy predictions. As such \(\mathbb{E}_\theta[H(p_\theta)] \approx 0\) and by \Cref{eq:mutualinformation}, this gives \(\mathrm{MI}(\bar{p};\theta) \approx H(\bar{p})\). Thus, MI closely tracks the entropy of the model average if there is no aleatoric uncertainty. Moreover, by Jensen’s inequality,
\[
-\log \bar{p}(y^\star) \;=\; -\log \mathbb{E}_\theta[p_\theta(y^\star)] \;\leq\; \mathbb{E}_\theta[-\log p_\theta(y^\star)],
\]
implying that if the individual models are accurate on average, the mean prediction is accurate as well. Therefore, we can apply \Cref{thm:low_entropy} again to conclude that low MI likely corresponds to low entropy in \(\bar{p}\), which in turn corresponds to low epistemic uncertainty (\Cref{fig:simplex} second-to-left). Again, the deterministic lower bound and probabilistic upper bound enforce a correlation between MI and the EU.

\paragraph{Takeaway}
For the regime of zero aleatoric uncertainty, there is a clear picture: consistent with empirical results of existing work, consistency-based and ensemble-based UQ are useful estimates of epistemic uncertainty. Our formal bounds establish a necessary correlation between these proxies for uncertainty and the true epistemic error of the model. This is particularly relevant for consistency-based estimators, which encapsulate many of the strongest and most widely-used UQ approaches \cite{vashurin-etal-2025-benchmarking}: Our results theoretically explain their empirical success if there is no aleatoric uncertainty.

\subsection{Current Uncertainty Quantification Fails Under Aleatoric Uncertainty}
\label{sec:non_trivial_au}
We now turn to the open question of how UQ methods perform under ambiguity in the answer distribution. 
We formally show \emph{why consistency-based estimators and ensembles fail once aleatoric uncertainty is present}.

\subsubsection{Consistency-Based Estimators Fail Under Aleatoric Uncertainty}
\label{subsec:theo_var_non_trivial}
In \Cref{subsec:theo_var_zero}, we establish two bounds that are valid under the assumption of zero aleatoric uncertainty:
\Cref{thm:high_entropy,thm:low_entropy} show that high predictive entropy implies high EU and vice versa. If their assumptions are violated this correlation is no longer enforced mathematically. 
High entropy in \(p^*\) may now reflect inherent ambiguity in the answer distribution that may be perfectly captured by the model that effectively has no epistemic error in this case.
(\Cref{fig:simplex} second-to-right).
Similarly, low entropy no longer implies epistemic confidence.
Intuitively, we can no longer assume that a well-trained model will naturally trend towards low-entropy predictions, as matching \(p^*\) may require predicting high-entropy p (\Cref{fig:simplex} right-most).

Together, this shows that, due to the absence of theoretical bounds that enforce a correlation, predictive entropy, and the wider field of consistency-based estimators that correlate strongly with entropy, are no longer reliable indicators of epistemic uncertainty under aleatoric uncertainty.

\newpage
Moreover, we can show that with just access to the predictive distribution \(p\), it is fundamentally impossible to disentangle EU and AU. This means that, under AU, proxies deriving EU from \(p\) must conflate both sources of uncertainty:

\begin{restatable}[Non-Identifiability of Epistemic Uncertainty]{proposition}{impossibility}
\label{prop:impossibility}
Let \(K \ge 2\) and \(\Delta^{K-1}\) be the probability simplex over $K$ classes. 
For any function \(f:\Delta^{K-1}\to\mathbb{R}\) and any \(p\in\Delta^{K-1}\), 
there exist \(p^*_1,p^*_{2}\in\Delta^{K-1}\) such that
\begin{align*}
\mathrm{KL}(p^*_1\!\parallel p)=0
\quad\text{and}\quad
\mathrm{KL}(p^*_2\!\parallel p) \geq \log K,
\end{align*}
Thus, the model's prediction \(p\), (and any function \(f(p)\)), can not distinguish between zero epistemic uncertainty (for $p_1^*$) and high epistemic uncertainty (\(\geq \log(K)\) for $p^*_2$). Consequently, such functions can not attribute predictive uncertainty to EU or AU reliably.
\end{restatable}

\subsubsection{Ensembles-based Estimators Fail Under Aleatoric Uncertainty}
\label{subsec:theo_ens_non_trivial}

Because of the strong dependence of mutual information as an estimator of EU and the entropy of the ensemble prediction \(\bar{p}\) (see \Cref{eq:mutualinformation}), \Cref{prop:impossibility} has immediate consequences for ensemble-based epistemic UQ as well.

\begin{restatable}[High MI $\not\Rightarrow$ High EU]{proposition}{trivialensemble}
\label{prop:nontrivial_au_ensembles}
Let \(K \ge 2\) and \(\Delta^{K-1}\) be the probability simplex over $K$ classes. Let \(\delta \in [0, \log K]\) be an arbitrary threshold on MI indicating uncertainty. Let \(p_\theta\) be such that \(\mathrm{MI}(\bar{p}; \theta) > \delta\) with \(\bar{p} = \mathbb{E}_\theta\left[p_\theta\right]\). Then $p^*=\bar{p} \in \Delta^{K-1}$ results in true epistemic uncertainty \(\mathrm{KL}(p^*\!\parallel \bar{p}) = 0\).
\end{restatable}

\Cref{prop:nontrivial_au_ensembles} stands in direct opposition to \Cref{subsec:theo_ens_zero}: In the zero AU case, high MI implied a high epistemic error in terms of the true \(p^*\) which was located in a corner of the probability simplex \(\Delta^{K-1}\). Lifting this restriction, for \emph{any} \(\bar{p}\) the true distribution $p^* = \bar{p}$ is associated with zero true EU no matter its associated MI (which is upper bounded by the entropy) (\Cref{fig:simplex} second-to-right). For instance, consider an ensemble where each member \(p_{\theta}\) assigns probability one to a distinct class. In this case, MI attains its maximum value, yet if \(p^*\) is uniform, the true EU is zero as the ensemble average perfectly recovers the uniform target. 
Hence, mutual information is likewise unreliable under ambiguity.

\paragraph{Takeaway} In contrast the positive results for estimating EU from consistency-based estimators and ensembles under zero AU, we formally show shortcomings of these estimators under AU. We reveal that both families of estimators (and conceptually similar proxies) can not properly attribute uncertainty to epistemic or aleatoric sources. This highlights a crucial gap in the literature on UQ for LLMs as answers with inherent ambiguity can not be overcome by the most prominent existing uncertainty estimation paradigms.

\section{A Novel Benchmark for Non-Zero Aleatoric Uncertainty}
\label{sec:dataset}

\begin{figure*}[t!]
    \centering

    \begin{subfigure}[t]{0.32\textwidth}
        \centering
        \includegraphics[width=\linewidth]{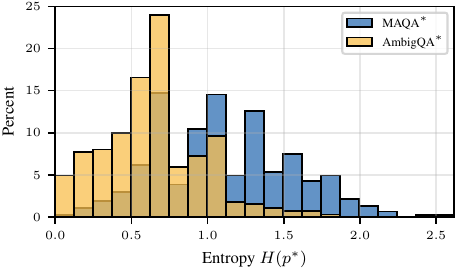}
        \captionsetup{labelformat=empty}
        \caption{\textit{Figure \thefigure a}: Distribution of ground-truth entropy \(H(p^*)\) across questions in MAQA\(^*\) and AmbigQA\(^*\).}
        \label{fig:entropy}
    \end{subfigure}
    \hfill
    \begin{subfigure}[t]{0.32\textwidth}
        \centering
        \includegraphics[width=\linewidth]{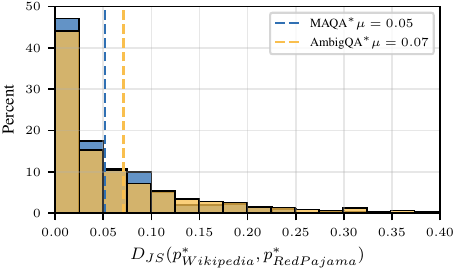}
        \captionsetup{labelformat=empty}
        \caption{\textit{Figure \thefigure b}: Distribution of JS divergences between using English Wikipedia and RedPjama-V1 for estimating \(p^*\).}
        \label{fig:js_divergence_main}
    \end{subfigure}
    \hfill
    \begin{subfigure}[t]{0.32\textwidth}
        \centering
        \includegraphics[width=\linewidth]{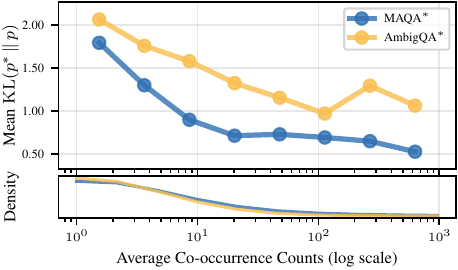}
        \captionsetup{labelformat=empty}
        \caption{\textit{Figure \thefigure c}: Model performance increases with co-occurrence, providing evidence for defining \(p^*\) based on co-occurrences.}
        \label{fig:cooccurence_performance}
    \end{subfigure}
    \label{fig:dataset_plots}
    \vspace{-1em}
\end{figure*}

Empirically verifying these results requires a dataset with ground-truth annotations \(p^*\).
In the absence of aleatoric uncertainty, this is straightforward as each question admits a single correct answer, and \(p^*\) corresponds to a one-hot distribution. In contrast, retrieving reliable ground-truth distributions under non-zero aleatoric uncertainty is notoriously challenging. However, factual QA admits a natural opportunity to obtain principled ground-truth distributions \(p^*\) from factual co-occurrence. We accordingly augment the ambiguous QA datasets MAQA \cite{yang2025maqaevaluatinguncertaintyquantification} and AmbigQA \cite{min-etal-2020-ambigqa} with such annotations. To our best knowledge, this is the first ambiguous QA datasets equipped with ground-truth probabilities. It enables principled UQ evaluation under ambiguity.

\subsection{Approximating  \(p^*\) from Corpus Statistics}
To approximate \(p^*\), we assume a frequentist's point of view: the probability of an outcome correlates with its relative frequency in the (pre-)training data. Concretely, for a question \(x\) and candidate answer \(y_i\), we approximate \(p^*(y_i \mid x)\) by the rate at which the underlying \emph{fact} occurs in the pre-training corpus. For example, if the statement ``Metformin is a medication for type 2 diabetes'' appears more often than ``Sulfonylureas is a medication for type 2 diabetes,'' then \(p^*(\text{Metformin}\mid x) \geq p^*(\text{Sulfonylureas}\mid x)\). This choice is well supported by previous results: Empirically, co-occurrence statistics correlate strongly with model performance as models score higher on samples with frequent co-occurrence \cite{kandpal2023largelanguagemodelsstruggle,mallen-etal-2023-trust}. Recently, \citet{wang2025generalization} demonstrated that, particularly in factual QA, LLM output probabilities correlate with these co-occurrence statistics. We also supply evidence by showing that increasing co-occurrence counts indicate better performance (\Cref{fig:cooccurence_performance}).
From a theoretical perspective, estimating these counts from statistics of \(p_{\text{train}}\) is better motivated than external training-agnostic annotations.
 As \(n \to \infty\), an ideal model converges to \(p_{\text{train}}\), and epistemic uncertainty vanishes entirely \citep{smith2025rethinking}. 
Such a reasoning implies that the reference distribution \(p^*\) should be realizable from the training data alone and thus must be dervied from the pretraining corpus.

\subsection{Obtaining the True Distribution \(p^*\)}
Since the pre-training datasets for LLMs are not publicly available, we instead employ the English Wikipedia \citet{structured-wikipedia} as a proxy for the pre-training corpus due to its widespread use in LLM pre-training and comprehensive coverage of factual knowledge. To perform the co-occurrence search, we use keywords extracted from the question alongside candidate answers (\Cref{tab:entailment_examples}). The keywords represent the most important words in the question, such as the question's subject. Importantly, both keywords and answers are stemmed to their base forms to ensure robustness against syntactic variation. \citet{elsahar-etal-2018-rex} demonstrate that subject–object co-occurrence is a reliable indicator for the presence of a subject–relation–object triplet, making it suitable for fact counting. We further improve the precision of these counts by using an entailment model to verify the factual occurrence of each candidate co-occurrence. The resulting datasets contain 468 and 2553 Q\&A examples, respectively. Their semantic answer-entropy distributions (\Cref{fig:entropy}) span a diverse range of annotated (aleatoric) distributions, with examples shown in \Cref{tab:dataset_examples}.

To validate the counts obtained through this method, we confirm their consistency with different co-occurrence counting strategies:
\begin{inparaenum}[(i)]
\item Utilizing keywords and answers but using as corpus the RedPajama-V1 dataset \citep{weber2024redpajama} via infini-gram \citep{liu2024infinigram}, and
\item through entity linking on the Pile dataset \citep{gao2020pile800gbdatasetdiverse} using DPBedia Spotlight \citep{kandpal2023largelanguagemodelsstruggle,isem2013daiber}. 
\end{inparaenum}
We find that the distributions obtained from all strategies align well, with Jensen–Shannon divergences between the estimated ground truth distributions \(p^*\) being negligibly small in most cases (\Cref{fig:js_divergence_main}). This validates the quality of our constructed ground-truth distributions \(p^*\) (also see \Cref{app:dataset_creation}).

\section{Experiments}
\label{sec:experiments}
\begin{table*}[t]
  \caption{Concordance scores \(AUC_c\) for all estimators on TriviaQA \((AU=0)\), MAQA\(^*\), and AmbigQA\(^*\) \(AU\geq0\).}
  \vspace{-0.2cm}
  \label{tab:auc_c_scores_all}
  \centering
  \resizebox{\textwidth}{!}{%
  \begin{tabular}{lcccccccccccccccccccccc}
    \toprule
    \multirow{4}{*}{\bfseries Model}
      & \multicolumn{7}{c}{\bfseries TriviaQA (\(AU=0\))}
      & \multicolumn{7}{c}{\bfseries MAQA\(^*\) (\(AU\geq0\))}
      & \multicolumn{7}{c}{\bfseries AmbigQA\(^*\) (\(AU\geq0\))} \\
    \cmidrule(lr){2-8}\cmidrule(lr){9-15}\cmidrule(lr){16-22}

      & \multicolumn{4}{c}{\bfseries Consistency}
      & \multicolumn{2}{c}{\bfseries Internal Rep.}
      & \multicolumn{1}{c}{\bfseries Ensemble}

      & \multicolumn{4}{c}{\bfseries Consistency}
      & \multicolumn{2}{c}{\bfseries Internal Rep.}
      & \multicolumn{1}{c}{\bfseries Ensemble}

      & \multicolumn{4}{c}{\bfseries Consistency}
      & \multicolumn{2}{c}{\bfseries Internal Rep.}
      & \multicolumn{1}{c}{\bfseries Ensemble} \\
    \cmidrule(lr){2-5}\cmidrule(lr){6-7}\cmidrule(lr){8-8}
    \cmidrule(lr){9-12}\cmidrule(lr){13-14}\cmidrule(lr){15-15}
    \cmidrule(lr){16-19}\cmidrule(lr){20-21}\cmidrule(lr){22-22}

      & \textbf{SE} & \textbf{MSP} & \textbf{SAR} & \textbf{IP}
      & \textbf{Linear} & \textbf{MLP} & \textbf{MI}

      & \textbf{SE} & \textbf{MSP} & \textbf{SAR} & \textbf{IP}
      & \textbf{Linear} & \textbf{MLP} & \textbf{MI}

      & \textbf{SE} & \textbf{MSP} & \textbf{SAR} & \textbf{IP}
      & \textbf{Linear} & \textbf{MLP} & \textbf{MI} \\
    \midrule

Llama 3.1-8B
& \cellcolor{rgb,255:red,197; green,170; blue,103}0.80
& \cellcolor{rgb,255:red,166; green,158; blue,119}0.74
& \cellcolor{rgb,255:red,189; green,167; blue,107}0.79
& \cellcolor{rgb,255:red,196; green,170; blue,104}0.80
& \cellcolor{rgb,255:red,125; green,142; blue,140}0.66
& \cellcolor{rgb,255:red,148; green,151; blue,128}0.71
& \cellcolor{rgb,255:red,219; green,179; blue,92}0.85

& \cellcolor{rgb,255:red,55; green,114; blue,175}0.52
& \cellcolor{rgb,255:red,48; green,112; blue,179}0.49
& \cellcolor{rgb,255:red,53; green,114; blue,176}0.51
& \cellcolor{rgb,255:red,60; green,116; blue,172}0.53
& \cellcolor{rgb,255:red,83; green,125; blue,161}0.57
& \cellcolor{rgb,255:red,84; green,126; blue,160}0.57
& \cellcolor{rgb,255:red,53; green,114; blue,176}0.51

& \cellcolor{rgb,255:red,100; green,132; blue,152}0.61
& \cellcolor{rgb,255:red,87; green,127; blue,159}0.58
& \cellcolor{rgb,255:red,99; green,132; blue,153}0.60
& \cellcolor{rgb,255:red,99; green,132; blue,153}0.60
& \cellcolor{rgb,255:red,81; green,125; blue,161}0.57
& \cellcolor{rgb,255:red,100; green,132; blue,152}0.61
& \cellcolor{rgb,255:red,85; green,126; blue,160}0.58 \\

Gemma 3-12B
& \cellcolor{rgb,255:red,249; green,191; blue,78}0.91
& \cellcolor{rgb,255:red,191; green,168; blue,106}0.79
& \cellcolor{rgb,255:red,222; green,180; blue,91}0.86
& \cellcolor{rgb,255:red,243; green,188; blue,80}0.90
& \cellcolor{rgb,255:red,126; green,142; blue,139}0.66
& \cellcolor{rgb,255:red,160; green,156; blue,122}0.73
& \cellcolor{rgb,255:red,219; green,179; blue,92}0.85

& \cellcolor{rgb,255:red,74; green,122; blue,165}0.55
& \cellcolor{rgb,255:red,63; green,118; blue,171}0.53
& \cellcolor{rgb,255:red,85; green,126; blue,160}0.58
& \cellcolor{rgb,255:red,96; green,131; blue,154}0.60
& \cellcolor{rgb,255:red,106; green,134; blue,149}0.62
& \cellcolor{rgb,255:red,92; green,129; blue,156}0.59
& \cellcolor{rgb,255:red,53; green,114; blue,176}0.51

& \cellcolor{rgb,255:red,124; green,142; blue,140}0.66
& \cellcolor{rgb,255:red,114; green,138; blue,145}0.64
& \cellcolor{rgb,255:red,124; green,142; blue,140}0.66
& \cellcolor{rgb,255:red,124; green,142; blue,140}0.66
& \cellcolor{rgb,255:red,75; green,122; blue,165}0.56
& \cellcolor{rgb,255:red,106; green,134; blue,149}0.62
& \cellcolor{rgb,255:red,85; green,126; blue,160}0.58 \\

Qwen 2.5-14B
& \cellcolor{rgb,255:red,231; green,184; blue,86}0.87
& \cellcolor{rgb,255:red,167; green,158; blue,119}0.74
& \cellcolor{rgb,255:red,204; green,173; blue,100}0.82
& \cellcolor{rgb,255:red,225; green,181; blue,89}0.86
& \cellcolor{rgb,255:red,121; green,140; blue,142}0.65
& \cellcolor{rgb,255:red,141; green,148; blue,131}0.69
& \cellcolor{rgb,255:red,219; green,179; blue,92}0.85

& \cellcolor{rgb,255:red,91; green,129; blue,157}0.59
& \cellcolor{rgb,255:red,75; green,122; blue,165}0.56
& \cellcolor{rgb,255:red,104; green,134; blue,150}0.62
& \cellcolor{rgb,255:red,90; green,128; blue,157}0.59
& \cellcolor{rgb,255:red,119; green,140; blue,142}0.65
& \cellcolor{rgb,255:red,115; green,138; blue,144}0.64
& \cellcolor{rgb,255:red,53; green,114; blue,176}0.51

& \cellcolor{rgb,255:red,133; green,145; blue,136}0.67
& \cellcolor{rgb,255:red,114; green,138; blue,145}0.63
& \cellcolor{rgb,255:red,131; green,144; blue,137}0.67
& \cellcolor{rgb,255:red,127; green,143; blue,138}0.66
& \cellcolor{rgb,255:red,92; green,129; blue,156}0.59
& \cellcolor{rgb,255:red,107; green,135; blue,148}0.62
& \cellcolor{rgb,255:red,85; green,126; blue,160}0.58 \\

\midrule

Llama 3.1-8B-I
& \cellcolor{rgb,255:red,212; green,176; blue,96}0.84
& \cellcolor{rgb,255:red,188; green,167; blue,108}0.79
& \cellcolor{rgb,255:red,208; green,175; blue,98}0.83
& \cellcolor{rgb,255:red,202; green,172; blue,101}0.81
& \cellcolor{rgb,255:red,135; green,146; blue,135}0.68
& \cellcolor{rgb,255:red,162; green,156; blue,121}0.73
& \cellcolor{rgb,255:red,231; green,184; blue,86}0.87

& \cellcolor{rgb,255:red,66; green,119; blue,169}0.54
& \cellcolor{rgb,255:red,57; green,115; blue,174}0.52
& \cellcolor{rgb,255:red,60; green,116; blue,172}0.53
& \cellcolor{rgb,255:red,66; green,119; blue,169}0.54
& \cellcolor{rgb,255:red,91; green,129; blue,157}0.59
& \cellcolor{rgb,255:red,79; green,124; blue,163}0.56
& \cellcolor{rgb,255:red,48; green,112; blue,178}0.50

& \cellcolor{rgb,255:red,96; green,131; blue,154}0.60
& \cellcolor{rgb,255:red,91; green,129; blue,157}0.59
& \cellcolor{rgb,255:red,97; green,131; blue,154}0.60
& \cellcolor{rgb,255:red,94; green,130; blue,155}0.60
& \cellcolor{rgb,255:red,84; green,126; blue,160}0.57
& \cellcolor{rgb,255:red,103; green,133; blue,151}0.61
& \cellcolor{rgb,255:red,96; green,130; blue,154}0.60 \\

Gemma 3-12B-I
& \cellcolor{rgb,255:red,174; green,161; blue,115}0.76
& \cellcolor{rgb,255:red,173; green,161; blue,116}0.76
& \cellcolor{rgb,255:red,181; green,164; blue,112}0.77
& \cellcolor{rgb,255:red,167; green,158; blue,119}0.74
& \cellcolor{rgb,255:red,154; green,153; blue,125}0.72
& \cellcolor{rgb,255:red,190; green,168; blue,107}0.79
& \cellcolor{rgb,255:red,231; green,184; blue,86}0.87

& \cellcolor{rgb,255:red,62; green,117; blue,171}0.53
& \cellcolor{rgb,255:red,66; green,119; blue,169}0.54
& \cellcolor{rgb,255:red,70; green,120; blue,167}0.55
& \cellcolor{rgb,255:red,66; green,119; blue,169}0.54
& \cellcolor{rgb,255:red,81; green,125; blue,162}0.57
& \cellcolor{rgb,255:red,77; green,123; blue,163}0.56
& \cellcolor{rgb,255:red,48; green,112; blue,178}0.50

& \cellcolor{rgb,255:red,84; green,126; blue,160}0.57
& \cellcolor{rgb,255:red,84; green,126; blue,160}0.57
& \cellcolor{rgb,255:red,85; green,126; blue,159}0.58
& \cellcolor{rgb,255:red,82; green,125; blue,161}0.57
& \cellcolor{rgb,255:red,82; green,125; blue,161}0.57
& \cellcolor{rgb,255:red,104; green,134; blue,150}0.62
& \cellcolor{rgb,255:red,96; green,130; blue,154}0.60 \\

Qwen 2.5-14B-I
& \cellcolor{rgb,255:red,159; green,155; blue,122}0.73
& \cellcolor{rgb,255:red,143; green,149; blue,131}0.69
& \cellcolor{rgb,255:red,161; green,156; blue,121}0.73
& \cellcolor{rgb,255:red,143; green,149; blue,131}0.69
& \cellcolor{rgb,255:red,169; green,159; blue,118}0.75
& \cellcolor{rgb,255:red,196; green,170; blue,104}0.80
& \cellcolor{rgb,255:red,231; green,184; blue,86}0.87

& \cellcolor{rgb,255:red,74; green,122; blue,165}0.55
& \cellcolor{rgb,255:red,69; green,120; blue,168}0.54
& \cellcolor{rgb,255:red,74; green,122; blue,165}0.56
& \cellcolor{rgb,255:red,71; green,121; blue,167}0.55
& \cellcolor{rgb,255:red,107; green,135; blue,149}0.62
& \cellcolor{rgb,255:red,97; green,131; blue,154}0.60
& \cellcolor{rgb,255:red,48; green,112; blue,178}0.50

& \cellcolor{rgb,255:red,84; green,126; blue,160}0.57
& \cellcolor{rgb,255:red,77; green,123; blue,163}0.56
& \cellcolor{rgb,255:red,83; green,125; blue,161}0.57
& \cellcolor{rgb,255:red,79; green,124; blue,163}0.56
& \cellcolor{rgb,255:red,87; green,127; blue,159}0.58
& \cellcolor{rgb,255:red,105; green,134; blue,150}0.62
& \cellcolor{rgb,255:red,96; green,130; blue,154}0.60 \\

\bottomrule
  \end{tabular}
  }
\end{table*}

Using this benchmark, we can empirically study the implications of our theoretical analysis of UQ and its relationship with aleatoric uncertainty in \Cref{sec:theory}.\footnote{Dataset and code available at
\href{https://hf.co/collections/ttomov/llm-uncertainty-under-ambiguity}{Hugging Face}
and
\href{https://github.com/timtomov/llm-uncertainty-under-ambiguity}{GitHub}.}

\subsection{Setup}

\textbf{Estimators}
We again categorize uncertainty estimators into the three previously described categories:
\begin{inparaenum}[(i)]
    \item \textbf{Consistency}: Semantic Entropy (SE) \citep{kuhn2023semanticuncertaintylinguisticinvariances}, Maximum Sentence Probability (MSP), and Shifting Attention to Relevance (SAR) \citep{duan2024shifting}. While not strictly in this category, we also test Iterative Prompting (IP) \citep{yadkori2024believebelievellm}, as it is specifically designed for cases of answer ambiguity.
    \item \textbf{Internal Representations:} We extract residual stream activations \(h^l\) at layer \(l\) for the final input token (pre-generation), and train linear probes and 2-layer MLPs with squared error loss to predict EU.
    \item \textbf{Ensembles:} We use an ensemble of three different LLMs to estimate Mutual Information (MI) \citep{depeweg2018decompositionuncertaintybayesiandeep}.
\end{inparaenum}

\textbf{Datasets}
For the zero AU setting, we use the single-answer QA dataset TriviaQA \cite{joshi-etal-2017-triviaqa}. For the non-zero AU case, we make use of our newly introduced MAQA\(^*\) and AmbigQA\(^*\) datasets.

\textbf{Models}
We evaluate the estimators across several models: LLaMA3.1 8B \citep{grattafiori2024llama3herdmodels}, Gemma3 12B \citep{gemmateam2025gemma3technicalreport}, Qwen2.5 14B \citep{qwen2025qwen25technicalreport}—each in both base and instruct variants. For ensembles, we combine these three architectures, treating them as approximate posterior samples from distinct model classes.

\textbf{Metrics}  
We study how well the estimated EU represents the true EU as quantified in \Cref{eq:decomposition}. Since both are continuous quantities, we use the \textbf{concordance} statistic \(AUC_c\) \citep{therneau2024concordance}. It quantifies the probability that the estimator correctly ranks a sample with a higher true EU above one with a lower true EU, \(\mathbb{P}(EU_i > EU_j \mid \text{Estimator}_i > \text{Estimator}_j)\). The resulting score can be interpreted analogously to the traditional AUC-ROC.

\subsection{Results}

\begin{figure}[t]
  \centering
  \includegraphics[width=\linewidth]{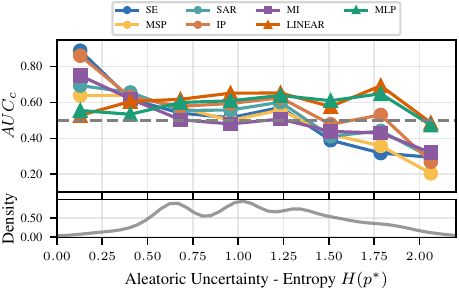}
    \caption{
\textbf{UQ performance versus aleatoric uncertainty}. The AUC decreases across all considered proxies at increasing aleatoric uncertainty. For highly ambiguous samples, many approaches are anti-correlated with true EU.
    }
\label{fig:entropy_performance}
\vspace{-1.5em}
\end{figure}

First, we establish that existing estimators perform well in the absence of aleatoric uncertainty consistent with existing work (\Cref{tab:auc_c_scores_all}, TriviaQA). With the exception of linear probes, all methods achieve strong performance in this regime. In contrast, when introducing aleatoric uncertainty, the performance collapses (\Cref{tab:auc_c_scores_all}, MAQA\(^*\) and AmbigQA\(^*\)). Methods that perform well in the zero AU setting achieve only marginal improvements over random chance. This deterioration is consistent across consistency-based, ensemble-based, and representation-based estimators. This aligns with our theoretical findings: In the absence of AU, the estimated uncertainty inevitably correlates well with the true epistemic error while there is little to no signal to estimate uncertainty on questions with ambiguous answers. The results for internal-state estimators indicate that they, too, can not capture information beyond what is described by the predictive distribution \(p\). We validate the robustness of these findings further in \Cref{app:predictive_var_exp,app:internal_rep_exp}.

\vspace{-1em}
\paragraph{Theoretical Bounds}

\begin{figure*}[t]
  \centering
  \includegraphics[width=\linewidth]{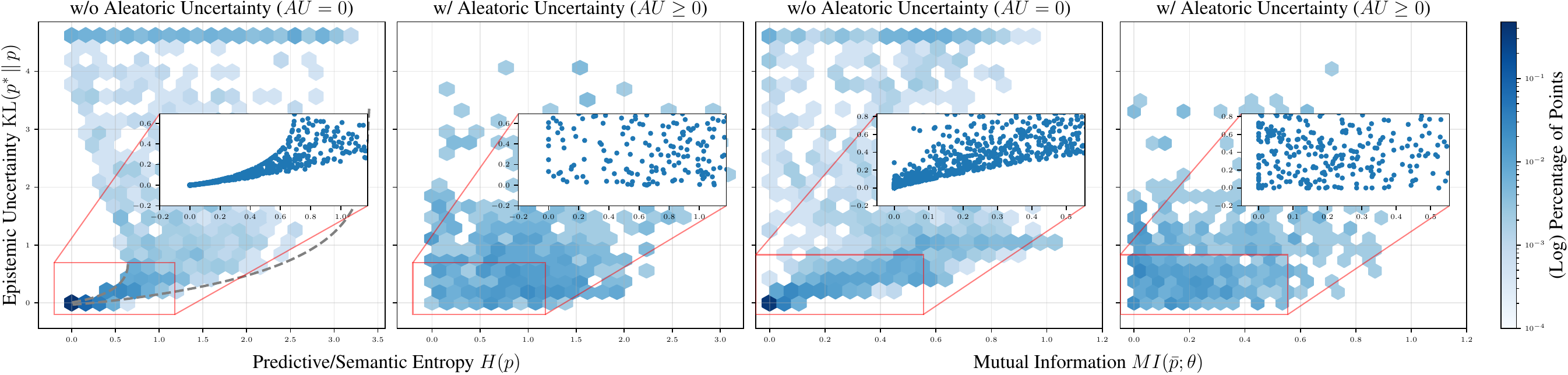}
  \caption{
   \textit{Relationship estimators and true epistemic uncertainty (EU) for Gemma 3-12B on zero aleatoric and non zero aleatoric data}.
    \textbf{Left: Predictive Entropy}: Under zero AU, predictive entropy is informative of EU due to lower and upper bounds in \Cref{subsec:theo_var_zero} (dashed gray lines). This correlation vanishes under non-trivial AU (\Cref{subsec:theo_var_non_trivial}).
     \textbf{Left: Mutual Information}: Under zero aleatoric uncertainty the bounds of \Cref{subsec:theo_ens_zero} lead to MI to be informative of EU. Again, this correlation vanishes under non-trivial AU (\Cref{subsec:theo_ens_non_trivial})
    }
    \vspace{-1em}
\label{fig:main_figure_results}
\end{figure*}

We empirically verify the bounds for the correlation of proxies for estimated and true EU for consistency-based models and ensembles under the absence of AU in \Cref{fig:main_figure_results}. The gray lines in the leftmost panel indicate the theoretical strict lower and probabilistic upper bound that force predictive entropy to align well with the epistemic uncertainty. In particular, we observe the lower bound to be tight and prevent high entropy predictions to correspond to low epistemic uncertainty. At the same time, the density of examples with low entropy and high epistemic uncertainty is low as per the probabilistic upper bound.
A similar trend arises for mutual information in ensemble-based UQ (third panel from the left).

Once AU is introduced, these theoretical bounds no longer hold and uncertainty proxies are no longer correlated with true EU. Predictive entropy and mutual information become unreliable indicators of epistemic uncertainty (second and fourth panel from the left). In particular, we observe pathological cases of low predicted EU and high true EU and vice versa for both entropy and mutual information.

\paragraph{Performance Deteriorates With Increasing Aleatoric Uncertainty}
We also study how UQ behaves under different degrees of AU besides just presence and absence thereof. As shown in \Cref{fig:entropy_performance}, UQ exhibits a clear downward trend as AU increases. In the extreme cases of high AU, estimators even negatively correlate with true EU. This pattern is most pronounced for consistency-based and ensemble-based methods.
As a consequence, the most useful interpretation of these UQ proxies depends on the amount of AU: If it is low, entropy indicates EU while under high AU it indicates confidence. Therefore, these estimators are unreliable in settings where we can not make assumptions about the underlying AU, aligning with the results in  \Cref{tab:auc_c_scores_all}.

\section{Related Work}
\label{sec:related_work}

\paragraph{UQ for LLMs}
A wide range of methods for uncertainty quantification in LLMs have been proposed \citep{vashurin-etal-2025-benchmarking,liu2025uncertaintyquantificationconfidencecalibration}. 
Of those, many approaches directly facilitate the model’s predictive distribution \(p\). 
The strongest empirical performance is typically achieved by \emph{consistency-based} methods \citep{vashurin-etal-2025-benchmarking}, which generate multiple outputs and assess their semantic variability \citep{kuhn2023semanticuncertaintylinguisticinvariances,nikitin2024kernellanguageentropyfinegrained,duan2024shifting}. 
Less computationally demanding approaches rely on a single output and estimate uncertainty from token-level or sentence-level probabilities, such as MSP. Another type of method assumes access to model internals. Prior work has shown that hidden representations can encode factual correctness \citep{li2023inferencetime,chen2024inside,orgad2025llms}, but has not examined settings with intrinsic ambiguity. Lastly, ensemble methods approximate posterior uncertainty by training multiple models and are often regarded as the gold standard in classical UQ \citep{lakshminarayanan2017simple}. However, due to their substantial computational cost, their application to LLMs is typically limited to fine-tuning \citep{balabanov2025uncertaintyquantificationfinetunedllms}.

\paragraph{Ambiguity in QA Tasks}
Prior work on uncertainty quantification in QA is predominantly benchmarked on datasets such as TriviaQA \citep{joshi-etal-2017-triviaqa}. These contain a single correct answer per question \citep{devic2025calibrationcollaborationllmuncertainty}. 
Consequently, most evaluations implicitly assume zero aleatoric uncertainty, and only few studies explicitly mention ambiguous settings. \citet{hou2024decomposinguncertaintylargelanguage,walha2025finegraineduncertaintydecompositionlarge} investigate aleatoric uncertainty that arises from ambiguous question phrasing and propose clarification-based methods to decompose epistemic and aleatoric uncertainty. However, this does not address settings in which ambiguity is inherent to the answer itself as studied in our work.
To address inherent answer ambiguity, \citet{yadkori2024believebelievellm} propose an estimator based on robustness to misleading contextual information. 
This approach relies on strong assumptions about LLM behavior that are unrealistic in practice, as shown by recent work on knowledge conflicts \citep{xie2024adaptivechameleonstubbornsloth,xu2024knowledgeconflictsllmssurvey}. 
In our experiments, we also find that this method is ineffective in ambiguous settings.

The limited evaluation of UQ methods under ambiguity reflects the lack of suitable benchmarks. Only few datasets explicitly target ambiguous QA, most notably AmbigQA \citep{min-etal-2020-ambigqa} and MAQA \citep{yang2025maqaevaluatinguncertaintyquantification}. 
To our knowledge, MAQA is the only dataset with inherent task ambiguity that cannot be resolved through more precise phrasing. However, the experimental setting of \citet{yang2025maqaevaluatinguncertaintyquantification} differs substantially from ours, as it requires the LLM to simultaneously generate all valid answers rather than approaching the problem from the lens of classification. 
FolkTexts \citep{cruz2024evaluatinglanguagemodelsrisk} provides an ambiguous QA dataset based on survey responses but lacks a ground-truth distribution \(p^*\). It is primarily used in calibration-style evaluations and not to study the disentanglement of uncertainty.

\section{Discussion}

\paragraph{Limitations}
Our new benchmark quantifies \(p^*\) as factual occurrences in Wikipedia. Although evidence suggests that occurrence frequency correlates well with model performance \citep{kandpal2023largelanguagemodelsstruggle,mallen-etal-2023-trust,wang2025generalization}, it has not been shown that LLMs approach this distribution in the infinite data limit. Besides aligning different co-occurrence counting strategies we also verify the robustness of our annotations with Dirichlet-distributed perturbations (\Cref{app:dirichlet}). Furthermore, our theoretical analysis encapsulates consistency-based and ensemble-based approaches while for internal representations our claims are only backed empirically accros all models. Additional experiments on classification probes support our conclusions further (see \Cref{app:internal_rep_exp}). Lastly, we follow existing work in phrasing UQ for LLMs as a classification problem \citep{kuhn2023semanticuncertaintylinguisticinvariances,aichberger2024rethinkinguncertaintyestimationnatural,aichberger2024many}. Scenarios that do not admit this perspective, e.g. simultaneously generating multiple responses, require a different framework.

\paragraph{The Role of Aleatoric Uncertainty in UQ for LLMs}

Our work is the first to provide a formal understanding of the relationship between UQ in LLMs and ambiguity. We establish bounds that force a correlation between uncertainty proxies and true epistemic uncertainty that only hold in the absence of aleatoric uncertainty. Further, we provide impossibility results for UQ using established estimators that explain why they are ineffective in the presence of aleatoric uncertainty. We back these results with an extensive empirical study that encompasses novel annotations for datasets with ambiguity.

\paragraph{Towards Reliable Uncertainty Estimators}  

Our works shows that established UQ paradigms (consistency, ensembles, internal representations) are unreliable once we can not assume zero aleatoric uncertainty. All of these methods are post-hoc approaches that attempt to extract UQ from LLMs that are not explicitly trained w.r.t. this objective. In standard classification problems, considering uncertainty has proven effective: For example, evidential deep learning \citep{sensoy2018evidentialdeeplearningquantify} learns second-order distributions over the predictive distribution to disentangle sources of uncertainty. Recent approaches train models on joint distributions to capture epistemic uncertainty \citep{johnson2024expertsdontcheatlearning,ahdritz2024provableuncertaintydecompositionhigherorder}. We hope that our theoretical and empirical results encourage a paradigm-shift toward actively modeling uncertainty during model training. Our results lay the groundwork for a principled and empirically measurable evaluation of UQ in LLMs that also considers ambiguity in the answers.

\section*{Acknowledgements}
The research presented has been performed in the frame of
the RADELN project funded by TUM Georg Nemetschek Institute Artificial Intelligence for the
Built World (GNI). It is further supported by the Bavarian Ministry of Economic Affairs, Regional
Development and Energy with funds from the Hightech Agenda Bayern.
\section*{Impact Statement}
In this work, we systematically analyze uncertainty quantification in Large Language Models. While any research may be misused, our primary goal is to improve the reliability of
these models to support their safe deployment in critical domains. We believe the benefits will
outweigh the potential risks.

\bibliography{references}

@misc{kuhn2023semanticuncertaintylinguisticinvariances,
	title        = {Semantic Uncertainty: Linguistic Invariances for Uncertainty Estimation in Natural Language Generation},
	author       = {Lorenz Kuhn and Yarin Gal and Sebastian Farquhar},
	year         = 2023,
	url          = {https://arxiv.org/abs/2302.09664},
	eprint       = {2302.09664},
	archiveprefix = {arXiv},
	primaryclass = {cs.CL}
}

@misc{walha2025finegraineduncertaintydecompositionlarge,
	title        = {Fine-Grained Uncertainty Decomposition in Large Language Models: A Spectral Approach},
	author       = {Nassim Walha and Sebastian G. Gruber and Thomas Decker and Yinchong Yang and Alireza Javanmardi and Eyke Hüllermeier and Florian Buettner},
	year         = 2025,
	url          = {https://arxiv.org/abs/2509.22272},
	eprint       = {2509.22272},
	archiveprefix = {arXiv},
	primaryclass = {cs.LG}
}

@misc{kirchhof2025positionuncertaintyquantificationneeds,
	title        = {Position: Uncertainty Quantification Needs Reassessment for Large-language Model Agents},
	author       = {Michael Kirchhof and Gjergji Kasneci and Enkelejda Kasneci},
	year         = 2025,
	url          = {https://arxiv.org/abs/2505.22655},
	eprint       = {2505.22655},
	archiveprefix = {arXiv},
	primaryclass = {cs.LG}
}

@misc{liu2025uncertaintyquantificationconfidencecalibration,
	title        = {Uncertainty Quantification and Confidence Calibration in Large Language Models: A Survey},
	author       = {Xiaoou Liu and Tiejin Chen and Longchao Da and Chacha Chen and Zhen Lin and Hua Wei},
	year         = 2025,
	url          = {https://arxiv.org/abs/2503.15850},
	eprint       = {2503.15850},
	archiveprefix = {arXiv},
	primaryclass = {cs.CL}
}

@misc{sensoy2018evidentialdeeplearningquantify,
	title        = {Evidential Deep Learning to Quantify Classification Uncertainty},
	author       = {Murat Sensoy and Lance Kaplan and Melih Kandemir},
	year         = 2018,
	url          = {https://arxiv.org/abs/1806.01768},
	eprint       = {1806.01768},
	archiveprefix = {arXiv},
	primaryclass = {cs.LG}
}

@misc{balabanov2025uncertaintyquantificationfinetunedllms,
	title        = {Uncertainty quantification in fine-tuned LLMs using LoRA ensembles},
	author       = {Oleksandr Balabanov and Hampus Linander},
	year         = 2025,
	url          = {https://arxiv.org/abs/2402.12264},
	eprint       = {2402.12264},
	archiveprefix = {arXiv},
	primaryclass = {cs.LG}
}

@misc{johnson2024expertsdontcheatlearning,
	title        = {Experts Don't Cheat: Learning What You Don't Know By Predicting Pairs},
	author       = {Daniel D. Johnson and Daniel Tarlow and David Duvenaud and Chris J. Maddison},
	year         = 2024,
	url          = {https://arxiv.org/abs/2402.08733},
	eprint       = {2402.08733},
	archiveprefix = {arXiv},
	primaryclass = {cs.LG}
}

@misc{xie2024adaptivechameleonstubbornsloth,
	title        = {Adaptive Chameleon or Stubborn Sloth: Revealing the Behavior of Large Language Models in Knowledge Conflicts},
	author       = {Jian Xie and Kai Zhang and Jiangjie Chen and Renze Lou and Yu Su},
	year         = 2024,
	url          = {https://arxiv.org/abs/2305.13300},
	eprint       = {2305.13300},
	archiveprefix = {arXiv},
	primaryclass = {cs.CL}
}

@misc{xu2024knowledgeconflictsllmssurvey,
	title        = {Knowledge Conflicts for LLMs: A Survey},
	author       = {Rongwu Xu and Zehan Qi and Zhijiang Guo and Cunxiang Wang and Hongru Wang and Yue Zhang and Wei Xu},
	year         = 2024,
	url          = {https://arxiv.org/abs/2403.08319},
	eprint       = {2403.08319},
	archiveprefix = {arXiv},
	primaryclass = {cs.CL}
}

@misc{ahdritz2024provableuncertaintydecompositionhigherorder,
	title        = {Provable Uncertainty Decomposition via Higher-Order Calibration},
	author       = {Gustaf Ahdritz and Aravind Gollakota and Parikshit Gopalan and Charlotte Peale and Udi Wieder},
	year         = 2024,
	url          = {https://arxiv.org/abs/2412.18808},
	eprint       = {2412.18808},
	archiveprefix = {arXiv},
	primaryclass = {cs.LG}
}

@misc{kotelevskii2025riskuncertaintygeneratingpredictive,
	title        = {From Risk to Uncertainty: Generating Predictive Uncertainty Measures via Bayesian Estimation},
	author       = {Nikita Kotelevskii and Vladimir Kondratyev and Martin Takáč and Éric Moulines and Maxim Panov},
	year         = 2025,
	url          = {https://arxiv.org/abs/2402.10727},
	eprint       = {2402.10727},
	archiveprefix = {arXiv},
	primaryclass = {stat.ML}
}

@misc{yang2025maqaevaluatinguncertaintyquantification,
	title        = {MAQA: Evaluating Uncertainty Quantification in LLMs Regarding Data Uncertainty},
	author       = {Yongjin Yang and Haneul Yoo and Hwaran Lee},
	year         = 2025,
	url          = {https://arxiv.org/abs/2408.06816},
	eprint       = {2408.06816},
	archiveprefix = {arXiv},
	primaryclass = {cs.AI}
}

@inproceedings{joshi-etal-2017-triviaqa,
	title        = {{T}rivia{QA}: A Large Scale Distantly Supervised Challenge Dataset for Reading Comprehension},
	author       = {Joshi, Mandar  and Choi, Eunsol  and Weld, Daniel  and Zettlemoyer, Luke},
	year         = 2017,
	month        = jul,
	booktitle    = {Proceedings of the 55th Annual Meeting of the Association for Computational Linguistics (Volume 1: Long Papers)},
	publisher    = {Association for Computational Linguistics},
	address      = {Vancouver, Canada},
	pages        = {1601--1611},
	doi          = {10.18653/v1/P17-1147},
	url          = {https://aclanthology.org/P17-1147/},
	editor       = {Barzilay, Regina  and Kan, Min-Yen}
}

@article{lakshminarayanan2017simple,
	title        = {Simple and scalable predictive uncertainty estimation using deep ensembles},
	author       = {Lakshminarayanan, Balaji and Pritzel, Alexander and Blundell, Charles},
	year         = 2017,
	journal      = {Advances in neural information processing systems},
	volume       = 30
}

@inproceedings{smith2025rethinking,
	title        = {Rethinking Aleatoric and Epistemic Uncertainty},
	author       = {Freddie Bickford Smith and Jannik Kossen and Eleanor Trollope and Mark van der Wilk and Adam Foster and Tom Rainforth},
	year         = 2025,
	booktitle    = {Forty-second International Conference on Machine Learning},
	url          = {https://openreview.net/forum?id=CY9MlORQs5}
}

@misc{nikitin2024kernellanguageentropyfinegrained,
	title        = {Kernel Language Entropy: Fine-grained Uncertainty Quantification for LLMs from Semantic Similarities},
	author       = {Alexander Nikitin and Jannik Kossen and Yarin Gal and Pekka Marttinen},
	year         = 2024,
	url          = {https://arxiv.org/abs/2405.20003},
	eprint       = {2405.20003},
	archiveprefix = {arXiv},
	primaryclass = {cs.LG}
}

@inproceedings{wang2025generalization,
	title        = {Generalization v.s. Memorization: Tracing Language Models{\textquoteright} Capabilities Back to Pretraining Data},
	author       = {Xinyi Wang and Antonis Antoniades and Yanai Elazar and Alfonso Amayuelas and Alon Albalak and Kexun Zhang and William Yang Wang},
	year         = 2025,
	booktitle    = {The Thirteenth International Conference on Learning Representations},
	url          = {https://openreview.net/forum?id=IQxBDLmVpT}
}

@misc{aichberger2024rethinkinguncertaintyestimationnatural,
	title        = {Rethinking Uncertainty Estimation in Natural Language Generation},
	author       = {Lukas Aichberger and Kajetan Schweighofer and Sepp Hochreiter},
	year         = 2024,
	url          = {https://arxiv.org/abs/2412.15176},
	eprint       = {2412.15176},
	archiveprefix = {arXiv},
	primaryclass = {cs.LG}
}

@misc{kandpal2023largelanguagemodelsstruggle,
	title        = {Large Language Models Struggle to Learn Long-Tail Knowledge},
	author       = {Nikhil Kandpal and Haikang Deng and Adam Roberts and Eric Wallace and Colin Raffel},
	year         = 2023,
	url          = {https://arxiv.org/abs/2211.08411},
	eprint       = {2211.08411},
	archiveprefix = {arXiv},
	primaryclass = {cs.CL}
}

@online{structured-wikipedia,
	title        = {Structured Wikipedia},
	author       = {Wikimedia Enterprise, Wikimedia Foundation},
	year         = 2024,
	month        = {sep}
}

@misc{hou2024decomposinguncertaintylargelanguage,
	title        = {Decomposing Uncertainty for Large Language Models through Input Clarification Ensembling},
	author       = {Bairu Hou and Yujian Liu and Kaizhi Qian and Jacob Andreas and Shiyu Chang and Yang Zhang},
	year         = 2024,
	url          = {https://arxiv.org/abs/2311.08718},
	eprint       = {2311.08718},
	archiveprefix = {arXiv},
	primaryclass = {cs.CL}
}

@misc{yadkori2024believebelievellm,
	title        = {To Believe or Not to Believe Your LLM},
	author       = {Yasin Abbasi Yadkori and Ilja Kuzborskij and András György and Csaba Szepesvári},
	year         = 2024,
	url          = {https://arxiv.org/abs/2406.02543},
	eprint       = {2406.02543},
	archiveprefix = {arXiv},
	primaryclass = {cs.LG}
}

@inproceedings{Lin_etal_SIGIR2021_Pyserini,
	title        = {{Pyserini}: A {Python} Toolkit for Reproducible Information Retrieval Research with Sparse and Dense Representations},
	author       = {Jimmy Lin and Xueguang Ma and Sheng-Chieh Lin and Jheng-Hong Yang and Ronak Pradeep and Rodrigo Nogueira},
	year         = 2021,
	booktitle    = {Proceedings of the 44th Annual International ACM SIGIR Conference on Research and Development in Information Retrieval (SIGIR 2021)},
	pages        = {2356--2362}
}

@article{vashurin-etal-2025-benchmarking,
	title        = {Benchmarking Uncertainty Quantification Methods for Large Language Models with {LM}-Polygraph},
	author       = {Vashurin, Roman  and Fadeeva, Ekaterina  and Vazhentsev, Artem  and Rvanova, Lyudmila  and Vasilev, Daniil  and Tsvigun, Akim  and Petrakov, Sergey  and Xing, Rui  and Sadallah, Abdelrahman  and Grishchenkov, Kirill  and Panchenko, Alexander  and Baldwin, Timothy  and Nakov, Preslav  and Panov, Maxim  and Shelmanov, Artem},
	year         = 2025,
	journal      = {Transactions of the Association for Computational Linguistics},
	publisher    = {MIT Press},
	address      = {Cambridge, MA},
	volume       = 13,
	pages        = {220--248},
	doi          = {10.1162/tacl_a_00737},
	url          = {https://aclanthology.org/2025.tacl-1.11/}
}

@misc{devic2025calibrationcollaborationllmuncertainty,
	title        = {From Calibration to Collaboration: LLM Uncertainty Quantification Should Be More Human-Centered},
	author       = {Siddartha Devic and Tejas Srinivasan and Jesse Thomason and Willie Neiswanger and Vatsal Sharan},
	year         = 2025,
	url          = {https://arxiv.org/abs/2506.07461},
	eprint       = {2506.07461},
	archiveprefix = {arXiv},
	primaryclass = {cs.CL}
}

@article{weber2024redpajama,
	title        = {RedPajama: an Open Dataset for Training Large Language Models},
	author       = {Maurice Weber and Daniel Y. Fu and Quentin Anthony and Yonatan Oren and Shane Adams and Anton Alexandrov and Xiaozhong Lyu and Huu Nguyen and Xiaozhe Yao and Virginia Adams and Ben Athiwaratkun and Rahul Chalamala and Kezhen Chen and Max Ryabinin and Tri Dao and Percy Liang and Christopher Ré and Irina Rish and Ce Zhang},
	year         = 2024,
	journal      = {NeurIPS Datasets and Benchmarks Track}
}

@misc{gao2020pile800gbdatasetdiverse,
	title        = {The Pile: An 800GB Dataset of Diverse Text for Language Modeling},
	author       = {Leo Gao and Stella Biderman and Sid Black and Laurence Golding and Travis Hoppe and Charles Foster and Jason Phang and Horace He and Anish Thite and Noa Nabeshima and Shawn Presser and Connor Leahy},
	year         = 2020,
	url          = {https://arxiv.org/abs/2101.00027},
	eprint       = {2101.00027},
	archiveprefix = {arXiv},
	primaryclass = {cs.CL}
}

@inproceedings{he2021deberta,
	title        = {DEBERTA: DECODING-ENHANCED BERT WITH DISENTANGLED ATTENTION},
	author       = {Pengcheng He and Xiaodong Liu and Jianfeng Gao and Weizhu Chen},
	year         = 2021,
	booktitle    = {International Conference on Learning Representations},
	url          = {https://openreview.net/forum?id=XPZIaotutsD}
}

@misc{depeweg2018decompositionuncertaintybayesiandeep,
	title        = {Decomposition of Uncertainty in Bayesian Deep Learning for Efficient and Risk-sensitive Learning},
	author       = {Stefan Depeweg and José Miguel Hernández-Lobato and Finale Doshi-Velez and Steffen Udluft},
	year         = 2018,
	url          = {https://arxiv.org/abs/1710.07283},
	eprint       = {1710.07283},
	archiveprefix = {arXiv},
	primaryclass = {stat.ML}
}

@inproceedings{liu2024infinigram,
	title        = {Infini-gram: Scaling Unbounded n-gram Language Models to a Trillion Tokens},
	author       = {Jiacheng Liu and Sewon Min and Luke Zettlemoyer and Yejin Choi and Hannaneh Hajishirzi},
	year         = 2024,
	booktitle    = {First Conference on Language Modeling},
	url          = {https://openreview.net/forum?id=u2vAyMeLMm}
}

@inproceedings{isem2013daiber,
	title        = {Improving Efficiency and Accuracy in Multilingual Entity Extraction},
	author       = {Joachim Daiber and Max Jakob and Chris Hokamp and Pablo N. Mendes},
	year         = 2013,
	booktitle    = {Proceedings of the 9th International Conference on Semantic Systems (I-Semantics)}
}

@misc{duan2024shifting,
	title        = {Shifting Attention to Relevance: Towards the Uncertainty Estimation of Large Language Models},
	author       = {Jinhao Duan and Hao Cheng and Shiqi Wang and Alex Zavalny and Chenan Wang and Renjing Xu and Bhavya Kailkhura and Kaidi Xu},
	year         = 2024,
	url          = {https://openreview.net/forum?id=yZJapMWdHZ}
}

@inproceedings{min-etal-2020-ambigqa,
	title        = {{A}mbig{QA}: Answering Ambiguous Open-domain Questions},
	author       = {Min, Sewon  and Michael, Julian  and Hajishirzi, Hannaneh  and Zettlemoyer, Luke},
	year         = 2020,
	month        = nov,
	booktitle    = {Proceedings of the 2020 Conference on Empirical Methods in Natural Language Processing (EMNLP)},
	publisher    = {Association for Computational Linguistics},
	address      = {Online},
	pages        = {5783--5797},
	doi          = {10.18653/v1/2020.emnlp-main.466},
	url          = {https://aclanthology.org/2020.emnlp-main.466/},
	editor       = {Webber, Bonnie  and Cohn, Trevor  and He, Yulan  and Liu, Yang}
}

@article{H_llermeier_2021,
	title        = {Aleatoric and epistemic uncertainty in machine learning: an introduction to concepts and methods},
	author       = {Hüllermeier, Eyke and Waegeman, Willem},
	year         = 2021,
	month        = mar,
	journal      = {Machine Learning},
	publisher    = {Springer Science and Business Media LLC},
	volume       = 110,
	number       = 3,
	pages        = {457–506},
	doi          = {10.1007/s10994-021-05946-3},
	issn         = {1573-0565},
	url          = {http://dx.doi.org/10.1007/s10994-021-05946-3}
}

@misc{qwen2025qwen25technicalreport,
	title        = {Qwen2.5 Technical Report},
	author       = {Qwen and : and An Yang and Baosong Yang and Beichen Zhang and Binyuan Hui and et al.},
	year         = 2025,
	url          = {https://arxiv.org/abs/2412.15115},
	eprint       = {2412.15115},
	archiveprefix = {arXiv},
	primaryclass = {cs.CL}
}

@misc{cruz2024evaluatinglanguagemodelsrisk,
	title        = {Evaluating language models as risk scores},
	author       = {André F. Cruz and Moritz Hardt and Celestine Mendler-Dünner},
	year         = 2024,
	url          = {https://arxiv.org/abs/2407.14614},
	eprint       = {2407.14614},
	archiveprefix = {arXiv},
	primaryclass = {cs.LG}
}

@misc{wu2025multiplereferencesmeaningfulvariations,
	title        = {Multiple References with Meaningful Variations Improve Literary Machine Translation},
	author       = {Si Wu and John Wieting and David A. Smith},
	year         = 2025,
	url          = {https://arxiv.org/abs/2412.18707},
	eprint       = {2412.18707},
	archiveprefix = {arXiv},
	primaryclass = {cs.CL}
}

@misc{maveli2025largelanguagemodelscapture,
	title        = {What can Large Language Models Capture about Code Functional Equivalence?},
	author       = {Nickil Maveli and Antonio Vergari and Shay B. Cohen},
	year         = 2025,
	url          = {https://arxiv.org/abs/2408.11081},
	eprint       = {2408.11081},
	archiveprefix = {arXiv},
	primaryclass = {cs.SE}
}

@misc{koupaee2025faithfulunfaithfulambiguousmultiagent,
	title        = {Faithful, Unfaithful or Ambiguous? Multi-Agent Debate with Initial Stance for Summary Evaluation},
	author       = {Mahnaz Koupaee and Jake W. Vincent and Saab Mansour and Igor Shalyminov and Han He and Hwanjun Song and Raphael Shu and Jianfeng He and Yi Nian and Amy Wing-mei Wong and Kyu J. Han and Hang Su},
	year         = 2025,
	url          = {https://arxiv.org/abs/2502.08514},
	eprint       = {2502.08514},
	archiveprefix = {arXiv},
	primaryclass = {cs.CL}
}

@misc{grattafiori2024llama3herdmodels,
	title        = {The Llama 3 Herd of Models},
	author       = {Aaron Grattafiori and Abhimanyu Dubey and Abhinav Jauhri and Abhinav Pandey and Abhishek Kadian et al.},
	year         = 2024,
	url          = {https://arxiv.org/abs/2407.21783},
	eprint       = {2407.21783},
	archiveprefix = {arXiv},
	primaryclass = {cs.AI}
}

@misc{gemmateam2025gemma3technicalreport,
	title        = {Gemma 3 Technical Report},
	author       = {Gemma Team and Aishwarya Kamath and Johan Ferret and Shreya Pathak and Nino Vieillard and et al.},
	year         = 2025,
	url          = {https://arxiv.org/abs/2503.19786},
	eprint       = {2503.19786},
	archiveprefix = {arXiv},
	primaryclass = {cs.CL}
}

@inproceedings{orgad2025llms,
	title        = {{LLM}s Know More Than They Show: On the Intrinsic Representation of {LLM} Hallucinations},
	author       = {Hadas Orgad and Michael Toker and Zorik Gekhman and Roi Reichart and Idan Szpektor and Hadas Kotek and Yonatan Belinkov},
	year         = 2025,
	booktitle    = {The Thirteenth International Conference on Learning Representations},
	url          = {https://openreview.net/forum?id=KRnsX5Em3W}
}

@inproceedings{li2023inferencetime,
	title        = {Inference-Time Intervention: Eliciting Truthful Answers from a Language Model},
	author       = {Kenneth Li and Oam Patel and Fernanda Vi{\'e}gas and Hanspeter Pfister and Martin Wattenberg},
	year         = 2023,
	booktitle    = {Thirty-seventh Conference on Neural Information Processing Systems},
	url          = {https://openreview.net/forum?id=aLLuYpn83y}
}

@misc{therneau2024concordance,
	title        = {Concordance},
	author       = {Terry M. Therneau and Elizabeth Atkinson},
	year         = 2024,
	month        = dec,
	url          = {https://cran.r-project.org/web/packages/survival/vignettes/concordance.pdf},
	note         = {Accessed: 2025-08-29},
	howpublished = {Vignette of the \texttt{survival} R package}
}

@inproceedings{chen2024inside,
	title        = {{INSIDE}: {LLM}s' Internal States Retain the Power of Hallucination Detection},
	author       = {Chao Chen and Kai Liu and Ze Chen and Yi Gu and Yue Wu and Mingyuan Tao and Zhihang Fu and Jieping Ye},
	year         = 2024,
	booktitle    = {The Twelfth International Conference on Learning Representations},
	url          = {https://openreview.net/forum?id=Zj12nzlQbz}
}

@inproceedings{mallen-etal-2023-trust,
	title        = {When Not to Trust Language Models: Investigating Effectiveness of Parametric and Non-Parametric Memories},
	author       = {Mallen, Alex  and Asai, Akari  and Zhong, Victor  and Das, Rajarshi  and Khashabi, Daniel  and Hajishirzi, Hannaneh},
	year         = 2023,
	month        = jul,
	booktitle    = {Proceedings of the 61st Annual Meeting of the Association for Computational Linguistics (Volume 1: Long Papers)},
	publisher    = {Association for Computational Linguistics},
	address      = {Toronto, Canada},
	pages        = {9802--9822},
	doi          = {10.18653/v1/2023.acl-long.546},
	url          = {https://aclanthology.org/2023.acl-long.546/},
	editor       = {Rogers, Anna  and Boyd-Graber, Jordan  and Okazaki, Naoaki}
}

@inproceedings{elsahar-etal-2018-rex,
	title        = {{T}-{RE}x: A Large Scale Alignment of Natural Language with Knowledge Base Triples},
	author       = {Elsahar, Hady  and Vougiouklis, Pavlos  and Remaci, Arslen  and Gravier, Christophe  and Hare, Jonathon  and Laforest, Frederique  and Simperl, Elena},
	year         = 2018,
	month        = may,
	booktitle    = {Proceedings of the Eleventh International Conference on Language Resources and Evaluation ({LREC} 2018)},
	publisher    = {European Language Resources Association (ELRA)},
	address      = {Miyazaki, Japan},
	url          = {https://aclanthology.org/L18-1544/},
	editor       = {Calzolari, Nicoletta  and Choukri, Khalid  and Cieri, Christopher  and Declerck, Thierry  and Goggi, Sara  and Hasida, Koiti  and Isahara, Hitoshi  and Maegaard, Bente  and Mariani, Joseph  and Mazo, H{\'e}l{\`e}ne  and Moreno, Asuncion  and Odijk, Jan  and Piperidis, Stelios  and Tokunaga, Takenobu}
}

@misc{gawlikowski2022surveyuncertaintydeepneural,
	title        = {A Survey of Uncertainty in Deep Neural Networks},
	author       = {Jakob Gawlikowski and Cedrique Rovile Njieutcheu Tassi and Mohsin Ali and Jongseok Lee and Matthias Humt and Jianxiang Feng and Anna Kruspe and Rudolph Triebel and Peter Jung and Ribana Roscher and Muhammad Shahzad and Wen Yang and Richard Bamler and Xiao Xiang Zhu},
	year         = 2022,
	url          = {https://arxiv.org/abs/2107.03342},
	eprint       = {2107.03342},
	archiveprefix = {arXiv},
	primaryclass = {cs.LG}
}

@misc{sanh2022multitaskpromptedtrainingenables,
	title        = {Multitask Prompted Training Enables Zero-Shot Task Generalization},
	author       = {Victor Sanh and Albert Webson and Colin Raffel and Stephen H. Bach and Lintang Sutawika and Zaid Alyafeai and Antoine Chaffin and Arnaud Stiegler and Teven Le Scao and Arun Raja and Manan Dey and M Saiful Bari and Canwen Xu and Urmish Thakker and Shanya Sharma Sharma and Eliza Szczechla and Taewoon Kim and Gunjan Chhablani and Nihal Nayak and Debajyoti Datta and Jonathan Chang and Mike Tian-Jian Jiang and Han Wang and Matteo Manica and Sheng Shen and Zheng Xin Yong and Harshit Pandey and Rachel Bawden and Thomas Wang and Trishala Neeraj and Jos Rozen and Abheesht Sharma and Andrea Santilli and Thibault Fevry and Jason Alan Fries and Ryan Teehan and Tali Bers and Stella Biderman and Leo Gao and Thomas Wolf and Alexander M. Rush},
	year         = 2022,
	url          = {https://arxiv.org/abs/2110.08207},
	eprint       = {2110.08207},
	archiveprefix = {arXiv},
	primaryclass = {cs.LG}
}

@inproceedings{aichberger2024many,
	title        = {How many opinions does your llm have? improving uncertainty estimation in nlg},
	author       = {Aichberger, Lukas and Schweighofer, Kajetan and Ielanskyi, Mykyta and Hochreiter, Sepp},
	year         = 2024,
	booktitle    = {ICLR 2024 Workshop on Secure and Trustworthy Large Language Models}
}
\bibliographystyle{icml2026}

\newpage

\crefalias{section}{appendix}
\crefalias{subsection}{appendix}
\crefalias{subsubsection}{appendix}

\appendix
\onecolumn

\section{Additional Experiments}
\subsection{Consistency Estimators}\label{app:predictive_var_exp}
\subsubsection{Accounting for Uncertainty in Estimating \(p^*\)}\label{app:dirichlet}
In practice, our estimate of the ground‐truth distribution \(p^*\) is itself uncertain due to limited or noisy co‐occurrence counts. To explicitly capture this uncertainty, we use a Dirichlet prior \(p^* \sim \mathrm{Dir}(\alpha)\), with parameters \(\alpha = (\alpha_1,\dots,\alpha_C)\). We start with a uniform prior \(\alpha_i = 1\) for all classes \(i\). After observing co‐occurrence counts \(n_i\), the posterior parameters become \(\alpha_i = 1 + n_i\). To prevent low‐count posteriors from remaining too uniform—which would erroneously decouple the model prediction \(p\) from \(p^*\)—we introduce a scaling factor \(\gamma\geq1\), defining  
\[
\alpha_i = 1 + \gamma\,n_i.
\]
Then, under the Dirichlet posterior, the \emph{aleatoric uncertainty} is given by:
\begin{align*}
\mathbb{E}_{p^*\sim\mathrm{Dir}(\alpha)}\bigl[H(p^*)\bigr] &= \mathbb{E}_{p^*\sim\mathrm{Dir}(\alpha)}\bigl[-\sum_{i=1}^Cp^*_ilog(p^*_i)\bigr]\\
&=-\sum_{i=1}^C\mathbb{E}_{p^*\sim\mathrm{Dir}(\alpha)}[p^*_ilog(p^*_i)]\\ &=-\sum_{i=1}^C[\frac{\alpha_i}{\alpha_0}(\psi(\alpha_i+1)-\psi(\alpha_0+1)]
\end{align*}
where \(\psi\) is the digamma function, and we leverage the fact that each \(p^*_i\sim Beta(\alpha_i,\alpha_0-\alpha_i)\). Likewise, the \emph{epistemic uncertainty} is defined as 
\begin{align*}
\mathbb{E}_{p^*\sim\mathrm{Dir}(\alpha)}\bigl[KL (p^* \,\|\, p)\bigr] &= \mathbb{E}_{p^*\sim\mathrm{Dir}(\alpha)}\bigl[CE(p^*||p)\bigr] - \mathbb{E}_{p^*\sim\mathrm{Dir}(\alpha)}\bigl[H(p^*)\bigr] \\
&=-\sum_{i=1}^C\mathbb{E}_{p^*\sim\mathrm{Dir}(\alpha)}[p_i^*]log(p_i) - \mathbb{E}_{p^*\sim\mathrm{Dir}(\alpha)}\bigl[H(p^*)\bigr] \\ 
&= -\sum_{i=1}^C\frac{\alpha_i}{\alpha_0}log(p_i) + \sum_{i=1}^C[\frac{\alpha_i}{\alpha_0}(\psi(\alpha_i+1)-\psi(\alpha_0+1) \\
&= \sum_{i=1}^C \frac{\alpha_i}{\alpha_0}\bigl[(\psi(\alpha_i+1)-\psi(\alpha_0+1)-log(p_i)\bigr]
\end{align*}

We perform ablation studies over different values of $\gamma$ (see \Cref{tab:dirchlet}). Increasing $\gamma$ corresponds to making a stronger assumption that the retrieved $p^*$ is exact, which causes the concordance score to approach the values reported in our main results. For smaller $\gamma$, $p^*$ becomes more independent of \(p\), especially given the relatively low counts noted earlier. Interestingly, estimator performance degrades further when we relax the assumption that \(p^*\) is exact, corroborating our main findings.

\begin{table}[h]
  \caption{Concordance scores $AUC_c$ for Gemma 3-12B for different likelihood multipliers ($\gamma$) across uncertainty estimators.}
  \label{tab:dirchlet}
  \centering
  \begin{small}
  \begin{tabular}{lcccccccc}
    \toprule
    \multirow{2}{*}{\bfseries Likelihood Multiplier \((\gamma)\)}
      & \multicolumn{4}{c}{\bfseries MAQA}
      & \multicolumn{4}{c}{\bfseries AmbigQA} \\
    \cmidrule(lr){2-5} \cmidrule(lr){6-9}
      & \textbf{SE} & \textbf{MSP} & \textbf{SAR} & \textbf{MI}
      & \textbf{SE} & \textbf{MSP} & \textbf{SAR} & \textbf{MI} \\
    \midrule
    $\gamma=1$ & \cellcolor{rgb,255:red,48; green,112; blue,179}0.50 & \cellcolor{rgb,255:red,48; green,112; blue,179}0.50 & \cellcolor{rgb,255:red,62; green,117; blue,171}0.53 & \cellcolor{rgb,255:red,69; green,120; blue,168}0.54 & \cellcolor{rgb,255:red,74; green,122; blue,165}0.56 & \cellcolor{rgb,255:red,79; green,124; blue,163}0.57 & \cellcolor{rgb,255:red,81; green,125; blue,162}0.57 & \cellcolor{rgb,255:red,75; green,122; blue,165}0.56 \\
    $\gamma=2$ & \cellcolor{rgb,255:red,53; green,114; blue,176}0.51 & \cellcolor{rgb,255:red,51; green,113; blue,177}0.51 & \cellcolor{rgb,255:red,67; green,119; blue,169}0.54 & \cellcolor{rgb,255:red,76; green,123; blue,164}0.56 & \cellcolor{rgb,255:red,86; green,127; blue,159}0.58 & \cellcolor{rgb,255:red,88; green,127; blue,158}0.58 & \cellcolor{rgb,255:red,91; green,129; blue,157}0.59 & \cellcolor{rgb,255:red,87; green,127; blue,159}0.58 \\
    $\gamma=5$ & \cellcolor{rgb,255:red,61; green,117; blue,172}0.53 & \cellcolor{rgb,255:red,57; green,115; blue,174}0.52 & \cellcolor{rgb,255:red,74; green,122; blue,165}0.55 & \cellcolor{rgb,255:red,84; green,126; blue,160}0.57 & \cellcolor{rgb,255:red,100; green,132; blue,152}0.61 & \cellcolor{rgb,255:red,97; green,131; blue,154}0.60 & \cellcolor{rgb,255:red,102; green,133; blue,151}0.61 & \cellcolor{rgb,255:red,100; green,132; blue,152}0.61 \\
    $\gamma=10$ & \cellcolor{rgb,255:red,65; green,118; blue,170}0.54 & \cellcolor{rgb,255:red,59; green,116; blue,173}0.52 & \cellcolor{rgb,255:red,77; green,123; blue,163}0.56 & \cellcolor{rgb,255:red,88; green,127; blue,158}0.58 & \cellcolor{rgb,255:red,107; green,135; blue,148}0.62 & \cellcolor{rgb,255:red,103; green,133; blue,151}0.61 & \cellcolor{rgb,255:red,109; green,136; blue,148}0.63 & \cellcolor{rgb,255:red,108; green,135; blue,148}0.62 \\
    $\gamma=100$ & \cellcolor{rgb,255:red,70; green,120; blue,167}0.55 & \cellcolor{rgb,255:red,62; green,117; blue,171}0.53 & \cellcolor{rgb,255:red,81; green,125; blue,161}0.57 & \cellcolor{rgb,255:red,94; green,130; blue,155}0.59 & \cellcolor{rgb,255:red,121; green,140; blue,142}0.65 & \cellcolor{rgb,255:red,112; green,137; blue,146}0.63 & \cellcolor{rgb,255:red,121; green,140; blue,142}0.65 & \cellcolor{rgb,255:red,121; green,140; blue,142}0.65 \\
    \midrule
    Real KL & \cellcolor{rgb,255:red,74; green,122; blue,165}0.55 & \cellcolor{rgb,255:red,63; green,118; blue,171}0.53 & \cellcolor{rgb,255:red,85; green,126; blue,160}0.58 & \cellcolor{rgb,255:red,96; green,131; blue,154}0.60 & \cellcolor{rgb,255:red,124; green,142; blue,140}0.66 & \cellcolor{rgb,255:red,114; green,138; blue,145}0.64 & \cellcolor{rgb,255:red,124; green,142; blue,140}0.66 & \cellcolor{rgb,255:red,124; green,142; blue,140}0.66 \\
    \bottomrule
  \end{tabular}
  \end{small}%
\end{table}

\subsubsection{Different \(p^*\) estimation methods}
\label{app:estimation_strategy}
We assess the robustness of our results by evaluating different strategies for estimating the ground-truth $p^*$, as outlined in \Cref{app:dataset_creation}. Across all estimators, the three methods yield highly similar results (\Cref{tab:p_star_estimation}), consistent with our observation that their estimated ground truths are strongly aligned. Note that, since we discard samples where at least one class has zero counts, different estimation strategies result in slightly different final datasets.

\begin{table}[h]
  \caption{Concordance scores $AUC_c$ for Gemma 3-12B for different estimation methods for ground truth\(p^*\)}
  \label{tab:p_star_estimation}
  \centering
  \begin{small}
  \begin{tabular}{lcccccccc}
    \toprule
    \multirow{2}{*}{\bfseries \(p^*\) Estimation Method}
      & \multicolumn{4}{c}{\bfseries MAQA}
      & \multicolumn{4}{c}{\bfseries AmbigQA} \\
    \cmidrule(lr){2-5} \cmidrule(lr){6-9}
      & \textbf{SE} & \textbf{MSP} & \textbf{SAR} & \textbf{MI}
      & \textbf{SE} & \textbf{MSP} & \textbf{SAR} & \textbf{MI} \\
    \midrule
    Wikipedia English & \cellcolor{rgb,255:red,74; green,122; blue,165}0.55 & \cellcolor{rgb,255:red,63; green,118; blue,171}0.53 & \cellcolor{rgb,255:red,85; green,126; blue,160}0.58 & \cellcolor{rgb,255:red,96; green,131; blue,154}0.60 & \cellcolor{rgb,255:red,124; green,142; blue,140}0.66 & \cellcolor{rgb,255:red,114; green,138; blue,145}0.64 & \cellcolor{rgb,255:red,124; green,142; blue,140}0.66 & \cellcolor{rgb,255:red,124; green,142; blue,140}0.66 \\
    RedPajama-V1 & \cellcolor{rgb,255:red,71; green,121; blue,167}0.55 & \cellcolor{rgb,255:red,59; green,116; blue,173}0.53 & \cellcolor{rgb,255:red,85; green,126; blue,160}0.58 & \cellcolor{rgb,255:red,93; green,129; blue,156}0.59 & \cellcolor{rgb,255:red,122; green,141; blue,141}0.65 & \cellcolor{rgb,255:red,111; green,137; blue,146}0.63 & \cellcolor{rgb,255:red,119; green,140; blue,142}0.65 & \cellcolor{rgb,255:red,122; green,141; blue,141}0.65 \\
    The Pile & \cellcolor{rgb,255:red,62; green,117; blue,171}0.53 & \cellcolor{rgb,255:red,63; green,118; blue,171}0.53 & \cellcolor{rgb,255:red,88; green,127; blue,158}0.58 & \cellcolor{rgb,255:red,88; green,127; blue,158}0.58 & \cellcolor{rgb,255:red,99; green,132; blue,153}0.60 & \cellcolor{rgb,255:red,85; green,126; blue,159}0.58 & \cellcolor{rgb,255:red,99; green,132; blue,153}0.60 & \cellcolor{rgb,255:red,100; green,132; blue,152}0.61 \\
    \bottomrule
  \end{tabular}
  \end{small}%
\end{table}

\subsubsection{Instruct Models Entropy Collapse}
\label{app:instruct}

\begin{figure}[t]
  \centering
  \includegraphics[width=\linewidth]{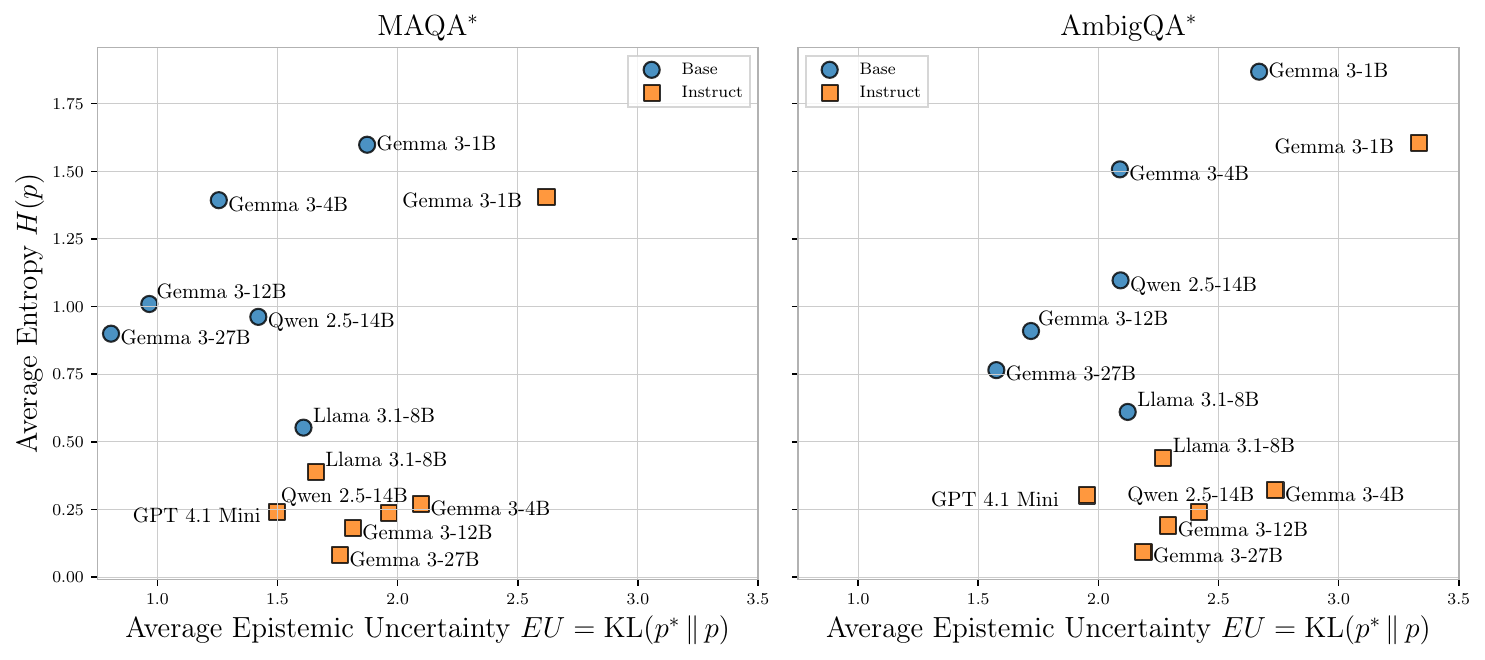}
  \caption{Entropy collapse of Instruct models on MAQA\(^*\) and AmbigQA\(^*\)}
  \label{fig:entropy_collapse}
\end{figure}

For instruct models, an additional insight is that the entropy for instruct models collapses to zero for most samples, even in cases with non-trivial aleatoric uncertainty. This behavior is undesirable, as it indicates that the models fail to represent any meaningful predictive distribution. Compared to base models, this collapse results in substantially worse model performance (average EU) (\Cref{fig:entropy_collapse}). Moreover, the entropy collapse also degrades estimator performance on TriviaQA, since a model that always outputs a single answer provides no variability and thus no basis to distinguish certain from uncertain cases.

\subsubsection{Effect of Model Size}
\label{app:model_size}
We evaluate different versions of Gemma~3—1B, 4B, 12B, and 27B—and observe that smaller models yield better performance for UQ estimation methods using variation of the predictive distribution (\Cref{tab:auc_c_scores_size}) on MAQA\(^*\). This effect appears to stem from the fact that smaller models often do not know the correct answers, and thus produce arbitrary outputs that form a high-entropy distribution. Such cases naturally coincide with high epistemic uncertainty, as the model lacks knowledge of the answers. Conversely, when a smaller model does know the answers, the resulting distribution has lower entropy and correspondingly lower epistemic uncertainty. As shown in \Cref{fig:entropy_collapse}, the average entropy decreases substantially with model size. Crucially, this reduction is accompanied by improved performance, indicating that larger models more accurately capture the underlying ground-truth distributions. The reduced estimator performance of smaller models on TriviaQA is consistent with prior observations in the literature \citep{kuhn2023semanticuncertaintylinguisticinvariances}.

\begin{table}[h]
  \caption{Concordance scores \(AUC_c\) for all estimators of different model sizes on TriviaQA (AU=0) and on AmbigQA\(^*\) \& MAQA\(^*\) (AU$\ge$0). An \(AUC_c = 0.50\) corresponds to random chance.}
  \label{tab:auc_c_scores_size}
  \centering
  \resizebox{\textwidth}{!}{%
  \begin{small}
  \begin{tabular}{lccccccccccccc}
    \toprule
    \multirow{3}{*}{\bfseries Model}
      & \multicolumn{4}{c}{\bfseries AU = 0}
      & \multicolumn{8}{c}{\bfseries AU $\ge$ 0} \\
    \cmidrule(lr){2-5}\cmidrule(lr){6-13}
      & \multicolumn{4}{c}{\bfseries TriviaQA}
      & \multicolumn{4}{c}{\bfseries MAQA}
      & \multicolumn{4}{c}{\bfseries AmbigQA} \\
    \cmidrule(lr){2-5}\cmidrule(lr){6-9}\cmidrule(lr){10-13}
      & \textbf{SE} & \textbf{MSP} & \textbf{SAR} & \textbf{MI}
      & \textbf{SE} & \textbf{MSP} & \textbf{SAR} & \textbf{MI}
      & \textbf{SE} & \textbf{MSP} & \textbf{SAR} & \textbf{MI} \\
    \midrule
    Gemma 3-1B & \cellcolor{rgb,255:red,187; green,166; blue,108}0.78 & \cellcolor{rgb,255:red,157; green,155; blue,123}0.72 & \cellcolor{rgb,255:red,179; green,163; blue,112}0.77 & \cellcolor{rgb,255:red,185; green,165; blue,110}0.78 & \cellcolor{rgb,255:red,142; green,149; blue,131}0.69 & \cellcolor{rgb,255:red,109; green,136; blue,148}0.63 & \cellcolor{rgb,255:red,152; green,152; blue,126}0.71 & \cellcolor{rgb,255:red,139; green,147; blue,133}0.69 & \cellcolor{rgb,255:red,129; green,143; blue,138}0.67 & \cellcolor{rgb,255:red,116; green,138; blue,144}0.64 & \cellcolor{rgb,255:red,115; green,138; blue,144}0.64 & \cellcolor{rgb,255:red,116; green,138; blue,144}0.64 \\
    Gemma 3-4B & \cellcolor{rgb,255:red,222; green,180; blue,91}0.85 & \cellcolor{rgb,255:red,177; green,162; blue,114}0.76 & \cellcolor{rgb,255:red,202; green,172; blue,101}0.82 & \cellcolor{rgb,255:red,218; green,178; blue,93}0.85 & \cellcolor{rgb,255:red,122; green,141; blue,141}0.65 & \cellcolor{rgb,255:red,81; green,125; blue,162}0.57 & \cellcolor{rgb,255:red,122; green,141; blue,141}0.65 & \cellcolor{rgb,255:red,130; green,144; blue,137}0.67 & \cellcolor{rgb,255:red,141; green,148; blue,132}0.69 & \cellcolor{rgb,255:red,121; green,140; blue,142}0.65 & \cellcolor{rgb,255:red,139; green,147; blue,133}0.69 & \cellcolor{rgb,255:red,133; green,145; blue,136}0.67 \\
    Gemma 3-12B & \cellcolor{rgb,255:red,249; green,191; blue,78}0.91 & \cellcolor{rgb,255:red,191; green,168; blue,106}0.79 & \cellcolor{rgb,255:red,222; green,180; blue,91}0.86 & \cellcolor{rgb,255:red,243; green,188; blue,80}0.90 & \cellcolor{rgb,255:red,74; green,122; blue,165}0.55 & \cellcolor{rgb,255:red,63; green,118; blue,171}0.53 & \cellcolor{rgb,255:red,85; green,126; blue,160}0.58 & \cellcolor{rgb,255:red,96; green,131; blue,154}0.60 & \cellcolor{rgb,255:red,124; green,142; blue,140}0.66 & \cellcolor{rgb,255:red,114; green,138; blue,145}0.64 & \cellcolor{rgb,255:red,124; green,142; blue,140}0.66 & \cellcolor{rgb,255:red,124; green,142; blue,140}0.66 \\
    Gemma 3-27B & \cellcolor{rgb,255:red,249; green,191; blue,78}0.93 & \cellcolor{rgb,255:red,194; green,169; blue,105}0.80 & \cellcolor{rgb,255:red,230; green,183; blue,87}0.87 & \cellcolor{rgb,255:red,249; green,191; blue,78}0.91 & \cellcolor{rgb,255:red,57; green,115; blue,174}0.52 & \cellcolor{rgb,255:red,57; green,115; blue,174}0.52 & \cellcolor{rgb,255:red,74; green,122; blue,165}0.56 & \cellcolor{rgb,255:red,83; green,125; blue,161}0.57 & \cellcolor{rgb,255:red,120; green,140; blue,142}0.65 & \cellcolor{rgb,255:red,107; green,135; blue,148}0.62 & \cellcolor{rgb,255:red,119; green,140; blue,142}0.65 & \cellcolor{rgb,255:red,118; green,139; blue,143}0.64 \\
    \bottomrule
  \end{tabular}
  \end{small}%
  }
\end{table}

\subsubsection{AUCROC for different uncertainty thresholds \(\delta\)}  
\label{app:binary_metrics}  
We also report \textbf{AUC-ROC}, where for a given threshold \(\delta\) we measure the separation between uncertain (\(EU \geq \delta\)) and certain (\(EU < \delta\)) samples. The results for different thresholds \(\delta\) are in (\Cref{tab:aucroc_scores_thresholds}).The higher values observed on AmbigQA\(^*\) are largely explained by its considerable proportion of near-zero entropy ground-truth samples.

\begin{table}[h]
  \caption{AUCROC scores for Gemma 3-12B for different uncertainty thresholds \(\delta\) across all estimators}
  \label{tab:aucroc_scores_thresholds}
  \centering
  \resizebox{\textwidth}{!}{%
  \begin{small}
  \begin{tabular}{lccccccccccccc}
    \toprule
    \multirow{3}{*}{\bfseries Uncertainty Threshold $\delta$}
      & \multicolumn{4}{c}{\bfseries AU = 0}
      & \multicolumn{8}{c}{\bfseries AU $\ge$ 0} \\
    \cmidrule(lr){2-5}\cmidrule(lr){6-13}
      & \multicolumn{4}{c}{\bfseries TriviaQA}
      & \multicolumn{4}{c}{\bfseries MAQA}
      & \multicolumn{4}{c}{\bfseries AmbigQA} \\
    \cmidrule(lr){2-5}\cmidrule(lr){6-9}\cmidrule(lr){10-13}
      & \textbf{SE} & \textbf{MSP} & \textbf{SAR} & \textbf{MI}
      & \textbf{SE} & \textbf{MSP} & \textbf{SAR} & \textbf{MI}
      & \textbf{SE} & \textbf{MSP} & \textbf{SAR} & \textbf{MI} \\
    \midrule
    $\delta = \log(1.5)$ & \cellcolor{rgb,255:red,249; green,191; blue,78}0.95 & \cellcolor{rgb,255:red,237; green,186; blue,83}0.89 & \cellcolor{rgb,255:red,249; green,191; blue,78}0.92 & \cellcolor{rgb,255:red,249; green,191; blue,78}0.94 & \cellcolor{rgb,255:red,70; green,120; blue,167}0.54 & \cellcolor{rgb,255:red,55; green,114; blue,175}0.52 & \cellcolor{rgb,255:red,85; green,126; blue,159}0.58 & \cellcolor{rgb,255:red,103; green,133; blue,151}0.61 & \cellcolor{rgb,255:red,185; green,165; blue,110}0.78 & \cellcolor{rgb,255:red,160; green,156; blue,122}0.73 & \cellcolor{rgb,255:red,179; green,163; blue,112}0.77 & \cellcolor{rgb,255:red,182; green,164; blue,111}0.78 \\
    $\delta = \log(2)$ & \cellcolor{rgb,255:red,249; green,191; blue,78}0.93 & \cellcolor{rgb,255:red,229; green,183; blue,87}0.87 & \cellcolor{rgb,255:red,246; green,190; blue,79}0.90 & \cellcolor{rgb,255:red,249; green,191; blue,78}0.92 & \cellcolor{rgb,255:red,75; green,122; blue,165}0.56 & \cellcolor{rgb,255:red,62; green,117; blue,171}0.53 & \cellcolor{rgb,255:red,96; green,131; blue,154}0.60 & \cellcolor{rgb,255:red,111; green,136; blue,147}0.63 & \cellcolor{rgb,255:red,170; green,160; blue,117}0.75 & \cellcolor{rgb,255:red,151; green,152; blue,127}0.71 & \cellcolor{rgb,255:red,165; green,158; blue,119}0.74 & \cellcolor{rgb,255:red,168; green,159; blue,118}0.75 \\
    $\delta = \log(3)$ & \cellcolor{rgb,255:red,244; green,189; blue,80}0.90 & \cellcolor{rgb,255:red,219; green,179; blue,92}0.85 & \cellcolor{rgb,255:red,239; green,187; blue,82}0.89 & \cellcolor{rgb,255:red,240; green,187; blue,82}0.89 & \cellcolor{rgb,255:red,88; green,128; blue,158}0.58 & \cellcolor{rgb,255:red,78; green,124; blue,163}0.56 & \cellcolor{rgb,255:red,106; green,134; blue,149}0.62 & \cellcolor{rgb,255:red,122; green,141; blue,141}0.65 & \cellcolor{rgb,255:red,161; green,156; blue,121}0.73 & \cellcolor{rgb,255:red,147; green,151; blue,129}0.70 & \cellcolor{rgb,255:red,159; green,155; blue,122}0.73 & \cellcolor{rgb,255:red,160; green,156; blue,122}0.73 \\
    \bottomrule
  \end{tabular}
  \end{small}%
  }
\end{table}

\subsection{Internal Representations}
\label{app:internal_rep_exp}

\begin{figure}[t]
  \centering
  \includegraphics[width=\linewidth]{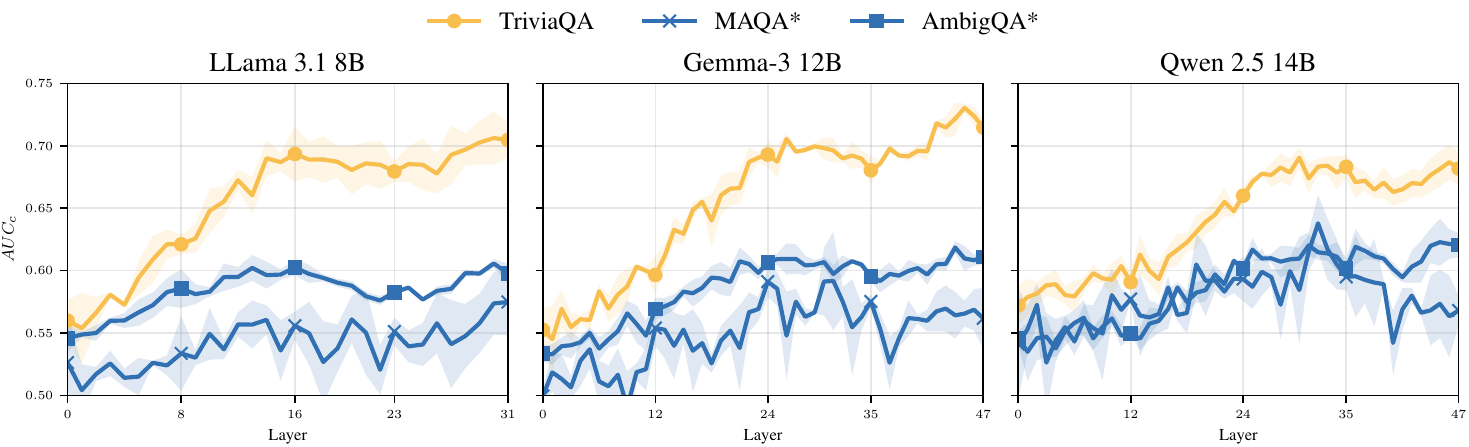}
    \caption{
    \textbf{MLP regression performance across layers}. Under zero AU, probes achieve satisfactory separation capability in deeper layers. Under non-trivial AU, performance collapses significantly, showing that hidden states do not reliably encode EU when ambiguity is present.}
\label{fig:mlp_probes_regression_full}
\end{figure}

\begin{figure}[t]
  \centering
  \includegraphics[width=\linewidth]{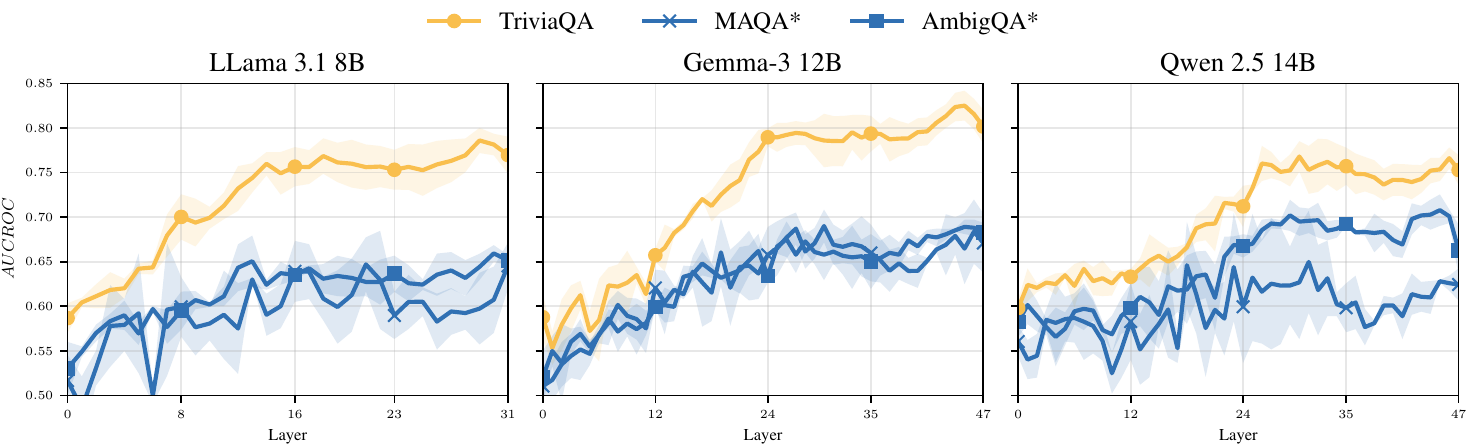}
    \caption{
    \textbf{MLP classification performance across layers}. Under zero AU, probes achieve satisfactory separation capability in deeper layers. Under non-trivial AU, performance collapses significantly, showing that hidden states do not reliably encode EU when ambiguity is present.}
\label{fig:probes_classification}
\end{figure}

In addition to our regression experiments \Cref{fig:mlp_probes_regression_full}, we train classifiers to predict a binary certainty label \(y \in \{0,1\}\). The label is obtained by thresholding the true epistemic uncertainty at \(\delta = \log(2)\), consistent with the procedure used in the previous experiments. We train linear probes \(\sigma(\langle \theta, h^l \rangle)\), where \(\sigma\) denotes the sigmoid function, and 2 Layer MLPs to distinguish between low and high epistemic uncertainty samples. \Cref{fig:probes_classification,tab:probes_auc_classification} shows the result for the MLP classification probes. Similarly, as in the regression case, we see a significant gap between performance on the different aleatoric uncertainty regimens.

\begin{table}[h]
  \caption{\(AUCROC\) for probes with certainty threshold \(\delta=\log(2)\).}
  \label{tab:probes_auc_classification}
  \centering
  \begin{small}
  \begin{tabular}{lcccccc}
    \toprule
    \multirow{3}{*}{\bfseries Model}
      & \multicolumn{2}{c}{\bfseries AU $= 0$}
      & \multicolumn{4}{c}{\bfseries AU $\ge 0$} \\
    \cmidrule(lr){2-3}\cmidrule(lr){4-7}
      & \multicolumn{2}{c}{\bfseries TriviaQA}
      & \multicolumn{2}{c}{\bfseries MAQA\(^*\)}
      & \multicolumn{2}{c}{\bfseries AmbigQA\(^*\)} \\
    \cmidrule(lr){2-3}\cmidrule(lr){4-5}\cmidrule(lr){6-7}
      & \textbf{Linear} & \textbf{MLP}
      & \textbf{Linear} & \textbf{MLP}
      & \textbf{Linear} & \textbf{MLP} \\
    \midrule
    Llama 3.1-8B & \cellcolor{rgb,255:red,174; green,161; blue,115}0.76 & \cellcolor{rgb,255:red,188; green,167; blue,108}0.79 & \cellcolor{rgb,255:red,114; green,138; blue,145}0.64 & \cellcolor{rgb,255:red,119; green,140; blue,142}0.65 & \cellcolor{rgb,255:red,114; green,138; blue,145}0.64 & \cellcolor{rgb,255:red,126; green,142; blue,139}0.66 \\
    Gemma 3-12B & \cellcolor{rgb,255:red,193; green,169; blue,106}0.80 & \cellcolor{rgb,255:red,208; green,174; blue,98}0.83 & \cellcolor{rgb,255:red,141; green,148; blue,131}0.69 & \cellcolor{rgb,255:red,141; green,148; blue,132}0.69  & \cellcolor{rgb,255:red,118; green,139; blue,143}0.64 & \cellcolor{rgb,255:red,141; green,148; blue,132}0.69 \\
    Qwen 2.5-14B & \cellcolor{rgb,255:red,167; green,158; blue,119}0.74 & \cellcolor{rgb,255:red,179; green,163; blue,112}0.77& \cellcolor{rgb,255:red,131; green,144; blue,137}0.67 & \cellcolor{rgb,255:red,121; green,140; blue,142}0.65 & \cellcolor{rgb,255:red,136; green,146; blue,134}0.68 & \cellcolor{rgb,255:red,149; green,151; blue,127}0.71  \\
    \bottomrule
  \end{tabular}
  \end{small}
\end{table}

\section{Implementation Details}
\label{app:implementation}

\subsection{Approximations}
\label{app:approx}
\paragraph{Approximation of \(p\)}
To estimate the probability \(p(y)\) of a semantic class \(y \in \mathcal{C}\), we sample \(K\) answers \(a_1, \dots, a_K\) from the model and then cluster them into semantic classes using an auxiliary entailment model. The probabilities of each semantic class are then obtained by aggregating and normalizing the answer probabilities within each class:
\[
p(y) \approx \frac{\tilde{p}(y)}{\sum_{j=1}^{|\mathcal{C}|} \tilde{p}(y_j)},
\quad \text{where} \quad
\tilde{p}(y) = \frac{1}{K} \sum_{i=1}^{K} \mathbb{I}(a_i \in y) p(a_i), \quad a_i \sim p(a).
\]
As \(K\rightarrow\infty\), the approximation converges to the model's true semantic answer distribution. We use a higher number of samples \(K=30\) to ensure a reasonable approximation. Semantic clustering follows the procedure of \cite{kuhn2023semanticuncertaintylinguisticinvariances}, employing a bi-directional entailment check with the \textit{deberta-v2-xlarge-mnli} model \cite{he2021deberta}. Samples are drawn via multinomial sampling with the default temperature, top-p, and top-k settings of each model. This choice is deliberate, as different model families and versions (e.g., base vs. instruct) provide different defaults, and we aim to evaluate them under their most realistic production settings.

\paragraph{Calculation of Epistemic Uncertainty \(KL(p^* \,\|\, p)\)}
The distribution \(p^*\) defines probabilities over the set of semantically distinct correct answers. Since the model distribution \(p\) is sampled and may be arbitrary, their supports need not coincide. Moreover, matching classes may also differ in surface form. As such, they need to be \emph{aligned} to be able to calculate the epistemic uncertainty. As an example, consider: 
\[
\begin{aligned}
p^* &= \{ \text{Heat}: 0.3, \text{Fuel}: 0.34, \text{Oxygen}: 0.36 \}\\
p   &= \{ \text{It's Heat}: 0.4, \text{Carbon}: 0.2, \text{Oxygen}: 0.4 \}.
\end{aligned}
\]
We construct a joint support set \(\{ \text{Heat}, \text{Fuel}, \text{Oxygen}, \text{Carbon} \}\), imputing missing values with \(0\) in \(p^*\) and with \(\epsilon=0.01\) in \(p\) to avoid undefined terms in the KL-divergence due to \(\log(0)\). Using \(\epsilon\) for the model distribution is justified, since in principle the model assigns non-zero probability to any possible sequence, making the support of \(p^*\) always a subset of the support of \(p\).  To determine the common support set, we apply the same semantic clustering procedure used for estimating \(p\), based on bidirectional entailment with \emph{deberta-v2-xlarge-mnli} \cite{he2021deberta}.

\subsection{Consistency Estimators}

\paragraph{Semantic Entropy (SE)}
For semantic entropy, we follow \citet{kuhn2023semanticuncertaintylinguisticinvariances} The method first estimates the semantic distribution $p$ as outlined in \Cref{app:approx} using \(K\) samples, and then computes the entropy:
\[
H(p) = - \sum_{i=1}^{|C|} p_i \log p_i.
\]

\paragraph{Maximum Sentence Probability (MSP)}
A simple yet effective estimator is the maximum sentence probability (MSP), defined as:
\[
\text{MSP} = 1 - \max_{a} \; p(a \mid x),
\]
where $p(a \mid x)$ is the probability assigned to answer $a$. Importantly, we do not compute $\max_y p(y \mid x)$ from the semantic distribution $p$ estimated above; instead, we directly perform beam search with 5 beams to identify the highest-probability answer. This approach is similar to a recent proposal by \citet{aichberger2024rethinkinguncertaintyestimationnatural}

\paragraph{Shifting Attention to Relevance (SAR)}
Instead of having hard clusters, SAR computes continuous semantic similarity scores to determine the importance of samples. Additionally, SAR mitigates the influence of irrelevant tokens by calculating the importance of each token on the semantics of the answer \citep{duan2024shifting}. We use the implementation of \citep{vashurin-etal-2025-benchmarking} using \emph{cross-encoder/stsb-roberta-large
} as the semantic similarity model and \(K=30\) samples.

\paragraph{Iterative Prompting (IP)}
The proposed estimator \citep{yadkori2024believebelievellm} should not be confused with the traditional MI estimator \citep{depeweg2018decompositionuncertaintybayesiandeep}. The core idea behind the method is based on the idea that if a model is epistemically certain, it is less likely to change its answer by the inclusion of a wrong answer in the input context. For a detailed explanation of this method, we refer to \citet{yadkori2024believebelievellm}. In our implementation, we limit the number of samples to \(K=10\). Conditional probabilities are obtained via teacher forcing and extracted explicitly from the model output. We use hyperparameters $\gamma_1=\gamma_2=10^{-9}$ and employ the prompt schema shown in Prompt~\ref{lst:decoy_prompt} to obtain the conditional probabilities.

\subsection{Internal Representation Estimators}
\paragraph{Activations}
We use the residual stream activations, evaluated at the final token position of the input sequence, i.e., immediately before the model begins generating the answer. This position captures the complete contextual representation of the question and is therefore a natural choice for probing. In our setting, answers are typically short, making the first generated token particularly important and further motivating this choice. We also experimented with probing MLP and attention activations, but observed no substantial differences.

\paragraph{Models} 
For linear baselines, we use ridge regression and logistic regression with default \texttt{scikit-learn} settings. For non-linear probes, we employ two-layer MLPs (hidden dimensions 256 and 128) with ReLU activations and the Adam optimizer, implemented via \texttt{scikit-learn}. 

\paragraph{Evaluation} 
All probes are evaluated with 3-fold cross-validation. In both regression and classification, we stratify the splits by binarized epistemic uncertainty (threshold \(\log(\delta)=2\)). Reported results are mean scores across folds, with standard deviations shown in the figures.

\subsection{Ensemble Estimator}  
As our ensemble-based estimator, we adopt the classical mutual information (MI) formulation \citep{depeweg2018decompositionuncertaintybayesiandeep}. Specifically, we treat LLaMA-3.1 8B, Gemma-3 12B, and Qwen-2.5 14B as approximate posterior samples from different architectures. The MI is then computed as the expected KL divergence between each member’s predictive distribution \(p_i\) and the ensemble mean \(\bar{p}\):  
\[
\mathrm{MI}(Y;\theta) \;=\; \frac{1}{3} \sum_{i=1}^{3} \mathrm{KL}\!\left(p_i \,\|\, \bar{p}\right),
\quad \text{where} \quad 
\bar{p} = \frac{1}{3}\sum_{i=1}^{3} p_i.
\]

As in the calculation of epistemic uncertainty, we align the distributions at the semantic level, following the exact procedure described in \Cref{app:approx}.

\subsection{Inference Prompts}
For base models, we employ few-shot prompts to guide the model toward producing answers in the desired format (Prompt~\ref{lst:base_prompt}). In contrast, instruct models are queried with a single instruction that specifies the expected answer style (Prompt~\ref{lst:instruct_prompt}).

\begin{lstlisting}[caption={Prompt for base models.},label={lst:base_prompt}]
Q: What is one planet in our solar system that has rings?
A: Saturn

Q: Name one programming language you know.
A: Python

Q: Who is one of the singers in the band ABBA?
A: Agnetha Faeltskog

Q: What is one color in the German flag?
A: Black

Q: {question}?
A:
\end{lstlisting}

\begin{lstlisting}[caption={Prompt for instruct models.},label={lst:instruct_prompt}]
Answer the following question with one word or phrase:
{question}?
\end{lstlisting}

\begin{lstlisting}[caption={Prompt for MI estimator},label={lst:decoy_prompt}]
A possible answer to the question {question} is {answer}.
Q: {question}?
A:
\end{lstlisting}

\section{Dataset Creation}
\label{app:dataset_creation}
Our dataset construction process consists of the following steps:
\begin{itemize}[label={}, leftmargin=*] 
\item \textbf{Question Rephrasing:} Each original question is reformulated to explicitly request exactly one specific answer. E.g.: \emph{"What are the essential components of the fire triangle?"} \(\rightarrow\)\emph{"What is one essential component of the fire triangle?"}. This prevents the model from producing multiple answers in a single generation. The rephrasing is done with \texttt{gpt-4.1-mini}.

\item \textbf{Keyword Extraction:} To enable the co-occurrence search, we extract a main keyword for the co-occurrence search. The keyword can either be a single word, like the subject, or a phrase. Critically, the co-occurrence of the keyword and the answer should reliably indicate the presence of the fact in the retrieved document. This is a valid assumption in most cases, as \cite{elsahar-etal-2018-rex} shows that when only the subject and object of a subject-object-relation triple co-occur in text, the resulting triple is often also present. However, for our main dataset \emph{Wikipedia English}, we take additional measures to enhance quality as explained \Cref{subsec:data_wikipedia}. The keyword extraction is done using \texttt{gpt-4.1-mini} with Prompt~\ref{lst:keyword_prompt} - except for the proxy using The Pile, which employs entity linking.

\item \textbf{Co-occurrence Search:} For each question, we perform a co-occurrence search for each answer on the proxy corpora. The final ground-truth distribution \(p^*(\cdot|q)\) for a given question \(q\) is then obtained by the relative frequency of the individual answer counts to all answer counts. To reduce potential biases, we discard samples in which at least one candidate answer has zero counts. Due to this fact, using the different proxies \emph{English Wikipedia}, \emph{RedPajama-V1}, and \emph{The Pile} can result in different samples in the final datasets.
\end{itemize}

\subsection{Wikipedia English}
\label{subsec:data_wikipedia}
\paragraph{Dataset curation}
We use the structured Wikipedia \cite{structured-wikipedia} dataset, and specifically the English version, which consists of all English article pages in a structured way. For each article, we are leveraging all data in the \emph{sections} tag. For the co-occurrence search, we use Pyserini and build the search index locally \cite{Lin_etal_SIGIR2021_Pyserini}. To define what constitutes a document—i.e., how articles are chunked for indexing—we leverage the dataset’s hierarchical structure: articles are organized into sections and subsections down to the level of individual paragraphs or lists. We assume that relevant facts are contained at this lowest level, which represents a coherent unit of text. The average length of the resulting chunks is around 65 words, with the distribution following a power-law: fewer than 1\% of the chunks exceed 300 words, while only a small number of outliers contain more than 2000 characters (\(\approx\)400 words). For such extreme outliers, we apply additional splitting at sentence boundaries. Importantly, apart from these rare cases, we keep the chunks intact and do not split them further, ensuring high recall of facts. Importantly, we also apply stemming to reduce words to their base forms, avoiding reliance on overly specific surface forms. The final index contains 65,069,586 documents. 
\paragraph{Co-occurrence counting}
In the retrieval step, we return all documents containing both the keyword and the candidate answer for a given question. 
Because the relationship between a question and its answer can be complex, relying on a single keyword often yields high recall but only moderate precision. For instance, consider the question \emph{``Who is the founder of Apple?''}---one valid answer is \emph{Steve Jobs}. 
If we extract \emph{Apple} as the main keyword, then any fact expressing \emph{``Steve Jobs founded Apple''} will naturally contain both \emph{Steve Jobs} and \emph{Apple}, which ensures high recall. 
However, the mere co-occurrence of \emph{Steve Jobs} and \emph{Apple} does not always capture the intended fact---for example, \emph{``Steve Jobs was the CEO of Apple''}. Such cases reduce precision. Hence, to ensure high precision, we apply an entailment procedure. Given a retrieved document through the co-occurrence search, we pass it to an LLM to verify that the fact is indeed present. For this step, we use \emph{ Gemma-3 12B Instruct} with the prompt shown in Prompt~\ref{lst:entailment_prompt} and examples in \Cref{tab:entailment_examples}. To keep the entailment step computationally feasible, we cap the number of retrieved documents per candidate answer at 1000—a threshold that we observe is rarely exceeded. The final number of samples for MAQA\(^*\) is 468 and for AmbigQA\(^*\) 2553 (\Cref{tab:pstar_stats}).

\begin{lstlisting}[caption={Prompt for entailment check.},label={lst:entailment_prompt}]
You are an expert at verifying factual entailment. I.e., is the fact 
present in the text? 
Given the following TEXT and FACT, answer with "yes" if the FACT 
follows from the TEXT, or "no" if it does not.

TEXT: {text}
FACT: {fact}
Answer:
\end{lstlisting}

\subsection{RedPajama-V1}
\paragraph{Dataset curation}
The Infini-Gram API provides access to co-occurrence counts across a range of large-scale pre-training datasets \cite{liu2024infinigram}. We use \emph{RedPajama-v1} \cite{weber2024redpajama}, which closely replicates the LLaMA pre-training corpus and includes a diverse set of data sources. 
\paragraph{Co-occurrence counting}
Similarly, as for Wikipedia English, we query for co-occurrences of the keyword with each candidate answer. For the Infini-Gram API we use the parameters \emph{max\_diff\_tokens}\(=100\) and \emph{max\_clause\_freq}\(=50000\). Since the underlying tokenizer (LLaMA 2) is sensitive to whitespaces for a keyword answer pair, we test all four combinations of including or removing a whitespace at the beginning of the keyword or answer. To obtain the final counts, we sum up the retrieved counts of the four different possibilities. Due to limited document access in Infini-Gram, we do not perform an entailment-checking phase. The final number of samples for MAQA\(^*\) is 470 and for AmbigQA\(^*\) 2331 (\Cref{tab:pstar_stats}).

\subsection{The Pile}
\paragraph{Dataset Curation}
In contrast to the previous two approaches, this method follows a different strategy for obtaining keywords and answers. It relies on entity linking, which identifies entities such as people, cities, or songs in both the question and the answer. The co-occurrence of a question entity with an answer entity is then retrieved from the Pile corpus \cite{gao2020pile800gbdatasetdiverse}. Following the approach of \cite{kandpal2023largelanguagemodelsstruggle}, we use the DBpedia Spotlight entity linker \cite{isem2013daiber} to extract entities from questions and answers. To improve accuracy, each answer is appended to its corresponding question before entity linking. When multiple candidate entities are returned for a question, we employ \emph{Gemma-3 12B Instruct} to filter for the most relevant one. The linker's parameters are set to \emph{confidence} \(=0.4\) and \emph{support} \(=1\).

\paragraph{Co-occurrence counting}
After obtaining the entity sets, we match them with pre-extracted entities from The Pile provided by \cite{kandpal2023largelanguagemodelsstruggle} to compute co-occurrence statistics. Similarly, as for RedPajama-V1, we do not perform an entailment-checking phase as we do not have access to the underlying documents.The final number of samples for MAQA\(^*\) is 120 and for AmbigQA\(^*\) 861 (\Cref{tab:pstar_stats}).

\begin{figure}[t]
  \centering
  \includegraphics[width=\linewidth]{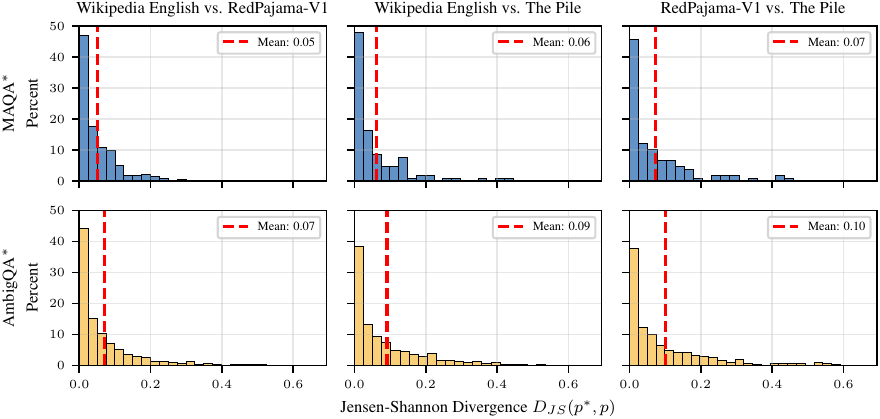}
  \caption{Comparison of retrieved ground-truth distribution \(p^*\) using different strategies}
  \label{fig:js_divergence}
\end{figure}

\subsection{Characteristics}
Summary statistics are reported in \Cref{tab:pstar_stats}. Compared to Wikipedia English, the other two strategies have access to a substantially larger pre-training corpus and therefore yield considerably higher counts. Nevertheless, the average entropies and their standard deviations remain in a similar range. As mentioned previously, we use \emph{English Wikipedia} as our principal strategy since it is the most controlled method with entailment checking, ensuring high precision and high recall. As can be seen in \Cref{tab:pstar_stats}, it also provides most samples on AmbigQA\(^*\) and similarly many on MAQA\(^*\) as RedPajama-V1. Using The Pile, in contrast, produces significantly fewer samples compared to the other two methods, as entity linking often can't find an entity in either question or answer, and hence such samples have to be discarded.
To assess how well the estimated ground truths \(p^*\) align across datasets, we compute the Jensen-Shannon divergence for all pairwise couplings on MAQA\(^*\) and AmbigQA\(^*\) (\Cref{fig:js_divergence}). The Jensen-Shannon divergence is given by: 
\(
\operatorname{JS}(p \,\|\, q)
\;=\;
\tfrac{1}{2}\,[\mathrm{KL}\!\left(p \,\middle\|\, m\right)
+
\,\mathrm{KL}\!\left(q \,\middle\|\, m\right)]
\) ,where \(
m = \tfrac{1}{2}(p+q)
\). It has the useful property of being symmetric, as we do not consider one strategy over the other as the truth. Overall, all strategies produce largely consistent ground truths, as reflected in the low average JS divergence and the characteristic power-law distribution (\Cref{fig:js_divergence}).

\begin{table}[h]
  \caption{Summary statistics for \(p^*\) estimation strategies: samples \(n\), mean answer-counts, and mean entropies (mean \(\pm\) std).}
  \label{tab:pstar_stats}
  \centering
  \begin{small}
  \resizebox{\textwidth}{!}{%
  \begin{tabular}{lcccccc}
    \toprule
    \multirow{2}{*}{\bfseries \(p^*\) Estimation Method}
      & \multicolumn{3}{c}{\bfseries MAQA\(^*\)} 
      & \multicolumn{3}{c}{\bfseries AmbigQA\(^*\)} \\
    \cmidrule(lr){2-4} \cmidrule(lr){5-7}
      & \textbf{n} 
      & \(\overline{\textbf{Count}}\pm\textbf{SD}\) 
      & \(\overline{H}\pm\textbf{SD}\)
      & \textbf{n} 
      & \(\overline{\textbf{Count}}\pm\textbf{SD}\) 
      & \(\overline{H}\pm\textbf{SD}\) \\
    \midrule
    Wikipedia English
      & 468 
      & 115.49 \(\pm\) 178.63 
      & 1.11 \(\pm\) 0.44 
      & 2553 
      & 55.81 \(\pm\) 104.77 
      & 0.64 \(\pm\) 0.33 \\
    RedPajama-V1
      & 470 
      & 143066.88 \(\pm\) 1234461.86 
      & 0.99 \(\pm\) 0.44 
      & 2331 
      & 1220812.64 \(\pm\) 29688717.59 
      & 0.52 \(\pm\) 0.35 \\
    The Pile
      & 120 
      & 46281.74 \(\pm\) 138909.80 
      & 0.97 \(\pm\) 0.44 
      & 861 
      & 19881.87 \(\pm\) 55441.02 
      & 0.60 \(\pm\) 0.33 \\
    \bottomrule
  \end{tabular}
  }%
  \end{small}
\end{table}

\section{Arbitrary Aleatoric Uncertainty}
\label{app:abitrary_au}
When constraining \(H(p^*)=0\), we implicitly restrict \(p^*\) to be a an indicator vector over one of the \(K\) classes. As shown in \Cref{thm:high_entropy,thm:low_entropy}, this setting allows, e.g, informative bounds on epistemic uncertainty using predictive entropy \(H(p)\). However, this is only one case. Consider instead the situation where \(p^*\) is known exactly. While this assumption is unrealistic (since complete knowledge of \(p^*\) renders estimation redundant), it helps illustrate that the issue is not the presence of aleatoric uncertainty but rather the fact that its extent is not known a priori. For example, if \(p^*\) is uniform, we obtain maximal aleatoric uncertainty with \(H(p^*)=\log K\). However, we can, in fact, exactly determine the epistemic uncertainty:
\[
EU=KL(p^*\mid\mid p)= \sum_{y}\frac{1}{K}\log(\frac{1}{Kp(y)})=-\log(K)-\frac{1}{K}\sum_{y}\log(p(y))
\]
Similarly when relaxing the constraint slightly to allow \(p^*\) be a high entropy distribution (e.g., \(H(p^*) \in [\log K - \epsilon, \log K]\)) estimation of epistemic uncertainty using \(H(p)\) should work reasonably: low predictive entropy necessarily implies high epistemic uncertainty, whereas high predictive entropy indicates lower epistemic uncertainty.

These illustrations clarify what we mean by \emph{arbitrary} aleatoric uncertainty: cases where no strong restrictions on \(H(p^*)\) are imposed. This is the typical regime in realistic applications, since constraining \(H(p^*)\) would require prior knowledge about the ambiguity structure of the task itself. This is especially the case in many linguistic problems, as a specific language task can have an arbitrary, ambiguous structure.

\section{Proofs}\label{app:proofs}

\impossibility*

\begin{proof}
Fix $p\in\Delta^{K-1}$. 
Set $p^*_1:=p$. Then $\mathrm{KL}(p^*_1\!\parallel p)=0$. 
Let $j \in \underset{i}{\arg\min}\quad p_i$ and define $p^*_2:=\mathbf{1}[y=j]$. Then
\[
\mathrm{KL}(p^*_2\!\parallel p) = -\log p_{\min}.
\]
Thus, for the same $p$, EU can be $0$ or large, while $f(p)$ is fixed. 
\end{proof}

\trivialensemble*

\begin{proof}
    Let there be a distribution $p_\theta$ such that $\mathrm{MI}(\bar{p}, \theta) > \delta$ for some $\delta \in [0, \log K]$. Since the probability simplex $\Delta^{K-1}$ is convex and $p_\theta \in \Delta^{K-1}$, the expected $\bar{p} = \mathbb{E}_\theta[p_\theta] \in \Delta^{K-1}$. Therefore, if the true distribution $p^* = \bar{p}$, the EU $\mathrm{KL}(p^* \| \bar{p})$ is trivially $0$. Thus, for any arbitrary estimate of EU through MI, there exists a true distribution with zero EU.
\end{proof}

\begin{proposition}[Zero aleatoric uncertainty implies EU is NLL]
\label{prop:zero_au}
\[
    H(p^*)= 0\implies EU = -\log(p(y=y^*))
\]
\end{proposition}

\begin{proof}
If \(H(p^*)=0\), then \(p^*(y)=\mathbf{1}[{y=y^*}]\). From this it follows:
\[
EU = KL(p^*\mid\mid p) = -\sum_{y\neq y^*}0\log(p(y)) - \log(p(y=y^*)) = -\log(p(y=y^*))
\]
\end{proof}

\thmhighentropy*
\begin{proof}
    We first define \(\alpha_{\delta}\) mathematically and how to obtain it.
    \[
    \alpha_\delta \;=\; \max \Bigl\{\, \max_{j} p_j \;:\; H(p) \ge \delta \,\Bigr\},
    \qquad \delta \in [0,\log K].
    \]
    Let \(H_{max}(\alpha)=-\alpha\log\alpha-(1-\alpha)\log\!\tfrac{1-\alpha}{K-1}\). This is the maximum entropy achievable by a distribution whose largest class probability is \(\alpha\in[1/K,1]\). Then \(\alpha_\delta\) is the solution of \(H_{max}(\alpha)=\delta\). Now we seek the lowest possible \(EU=-log(p(y=y^*))\) under the constraint $H(p) \geq \delta$. This exactly occurs if the maximal possible probability \(\alpha_\delta\) is on the correct class and hence \(EU=-\log(p(y=y^*))\geq-\log(\alpha_{\delta})\)
\end{proof}

\thmlowentropy*
\begin{proof}
We first define \(\gamma_{\delta}\) mathematically and how to obtain it:
\[
\gamma_\delta \;=\; \min \Bigl\{\, \max_{j} p_j \;:\; H(p) \le \delta \,\Bigr\},
\qquad \delta \in [0,\log 2],
\]
Denote \(H_{B}(\gamma)=-\gamma\log\gamma-(1-\gamma)\log(1-\gamma)\) as the binary entropy function. Then \(\gamma_{\delta}\) is the solution of \(H_B(\gamma) =\delta\) for \(\gamma\in[1/2,1]\) and we can now proceed:
\begin{align}
    \mathcal{L} &= \mathbb{E}_{(x,y^*)}\bigl[-\log p_{y^*}\bigr] \\ 
    &= \mathbb{E}_{(x,y^*)}\bigl[-\log p_{y^*}\mid H(p) \leq \delta \bigr] \mathbb{P}(H(p)\leq \delta)\label{eq:iterativelaw1}\\ &\quad + \mathbb{E}_{(x,y^*)}\bigl[-\log p_{y^*}\mid H(p) > \delta \bigr] \mathbb{P}(H(p)>\delta) \notag \\
    &=  \mathbb{E}_{(x,y^*)}\bigl[-\log p_{y^*}\mid H(p) \leq \delta \cap \arg\max p \neq y^* \bigr] \mathbb{P}(H(p)\leq \delta \cap \arg\max p \neq y^* )\label{eq:iterativelaw2}\\ &\quad+
    \mathbb{E}_{(x,y^*)}\bigl[-\log p_{y^*}\mid H(p) \leq \delta \cap \arg\max p = y^* \bigr] \mathbb{P}(H(p)\leq \delta \cap \arg\max p = y^*)\notag\\ &\quad+ 
\mathbb{E}_{(x,y^*)}\bigl[-\log p_{y^*}\mid H(p) > \delta \bigr] \mathbb{P}(H(p)>\delta)\notag\\
&\geq -\log(1-\gamma_{\delta})\mathbb{P}(H(p)\leq \delta \cap \arg\max p \neq y^*) -\log(\alpha_\delta)\mathbb{P}(H(p)>\delta)\label{eq:bound}
\end{align}
Where we use in \ref{eq:iterativelaw1} the law of total expectation to separate into high and low entropy predictions. In \ref{eq:iterativelaw2}, we further partition the space of low entropy predictions into correct and incorrect ones. Lastly, in \ref{eq:bound}, we bound the expectation values. High entropy predictions occur at least loss \(-\log(\alpha_{\delta})\) according to \cref{thm:high_entropy}. Low entropy predictions that are incorrect will have maximally \(1-\gamma_{\delta}\) mass on an \emph{correct} class and as such occur at least \(-\log(1-\gamma_{\delta})\) loss.
Rearranging terms and substituting \(\mathbb{P}(H(p)> \delta)=1-\mathbb{P}(H(p)\leq \delta)\)yields
\[
\mathbb{P}(H(p)\leq \delta \cap\arg\max p \neq y^*)\leq \frac{\mathcal{L}+(1-\mathbb{P}(H(p)\leq \delta))\log(\alpha_\delta)}{-\log(1-\gamma_{\delta})}
\]
Dividing by \(\mathbb{P}(H(p)\leq\delta)\) we finally get the conditional bound:
\[
\mathbb{P}(\arg\max p \neq y^*\mid H(p) \leq \delta) \leq\frac{\mathcal{L}+(1-\mathbb{P}(H(p)\leq \delta))\log(\alpha_\delta)}{-\log(1-\gamma_{\delta})\mathbb{P}(H(p)\leq\delta)}
\]
which can be rewritten to obtain an upper bound as:
\begin{align}
\mathbb{P}(\arg\max p =y^* \mid H(p) \leq \delta) &\geq 1-\frac{\mathcal{L}+(1-\mathbb{P}(H(p)\leq \delta))\log(\alpha_\delta)}{-\log(1-\gamma_{\delta})\mathbb{P}(H(p)\leq\delta)}\\
&= 1 - \frac{\mathcal{L}}{-\log(1-\gamma_{\delta})\mathbb{P}(H(p)\leq\delta)}\\&+\frac{-\log(\alpha_\delta)(1-\mathbb{P}(H(p)\leq \delta))}{-\log(1-\gamma_{\delta})\mathbb{P}(H(p)\leq\delta)}
\end{align}

Realizing that \(-\log (p_{y^*}) \leq-\log(\gamma_{\delta} )\;\Longleftrightarrow\; \arg\max p = y^*\) - since \(\gamma_{\delta}\) is the minimum possible maximum probability - we get:
\begin{align}
\mathbb{P}(\log p_{y^*} \leq-\log(\gamma_{\delta})\mid H(p) \leq \delta) &\geq 1 - \frac{\mathcal{L}}{-\log(1-\gamma_{\delta})\mathbb{P}(H(p)\leq\delta)}\\&+\frac{-\log(\alpha_\delta)(1-\mathbb{P}(H(p)\leq \delta))}{-\log(1-\gamma_{\delta})\mathbb{P}(H(p)\leq\delta)}
\end{align}

Abbreviating \(-\log p_{y^*}\) as \emph{epistemic uncertainty} EU and simplifying by leaving out the second term, we obtain the bound stated in the theorem.
\begin{align}
\mathbb{P}(EU\leq-\log(\gamma_{\delta})\mid H(p) \leq \delta) &\geq 1 - \frac{\mathcal{L}}{-\log(1-\gamma_{\delta})*\mathbb{P}(H(p)\leq\delta)}
\end{align}

\end{proof}

\clearpage
\begin{table}[t]
  \caption{Examples of entailment check in the co-occurrence pipeline for Wikipedia English}
  \label{tab:entailment_examples}
  \centering
  \resizebox{\textwidth}{!}{%
    \begin{tabular}{l p{3.5cm} p{2.5cm} p{2cm} p{4.5cm} p{4.5cm}}
      \toprule
      \textbf{Idx} & \textbf{Question} & \textbf{Keyword} & \textbf{Answer} & \textbf{Positive Example} & \textbf{Negative Example} \\
      \midrule
72 & Who was the recipient of the Bharat Ratna award when it was first awarded? & Bharat Ratna & ['C. Rajagopalachari'] & article rajaji national park, section abstract:
rajaji national park was named after c. rajagopalachari (rajaji), a prominent leader of the freedom struggle, the first and last governor-general of independent india and one of the first recipients of india's highest civilian award, bharat ratna (in 1954). & article central college, bangalore, section notable students:
bharat ratna sir m. visvesvaraya, harshavardhan mudaliar, prof in english, bharat ratna c. rajagopalachari, bharat ratna c. n. r. rao, indian chemist, shivakumara swami, pusapati vijayarama gajapati raju, maharaja of vizianagaram, h. narasimhaiah, guruswami mudaliar, hospet sumitra, n. santosh hegde, justice, navaratna rama rao, leading administrator, author and founder of the sericulture department, n. s. subba rao, maya rao (1928-20... \\
      \midrule
186 & What is one specific type of agricultural product the Wachau Valley is known for? & Wachau Valley & ['grapes'] & article wachau, section wine:
the wachau valley is well known for its production of apricots and grapes, both of which are used to produce specialty liquors and wines. the wine district's rolling vineyards produce complex white wines. wachau is a source of austria's most prized dry rieslings and grüner veltliners, some of the best from the steep stony slopes next to the danube on which the vines are planted. the temperature variation in the valley between day and cold nights has a significant ro... & No negative example found \\
      \midrule
198 & What is the name of one Unforgivable Curse from the Harry Potter books? & Unforgivable Curses & ['Imperio'] & article imperio, section abstract:
imperio, a curse in the harry potter series (see magic in harry potter\#unforgivable curses), imperio (band), austrian band & No negative example found \\
      \midrule
205 & Who was one of the main cast members of 'The Big Valley' TV show? & The Big Valley & ['Barbara Stanwyck'] & article the big valley, section reception : popularity:
in the comedy film airplane! (1980), the wacky air traffic controller johnny, played by stephen stucker, paid homage to big valley ' s penchant for big drama in one of his many asides. after lloyd bridges ' character frets about a pilot who cracked under pressure, johnny says: "it happened to barbara stanwyck!" and "nick, heath, jarrod – there's a fire in the barn!" the big valley also has seeped into the darker cinematic subconscious. in b... & article peter breck, section career : after the big valley:
on january 20, 1990, while teaching at the drama school, breck was notified of barbara stanwyck's death. she requested no funeral nor memorial. \\
      \midrule
297 & Which stadium did the New Orleans Saints use for their home games in the seasons following Hurricane Katrina in 2005? & New Orleans Saints & ['Alamodome'] & article tom benson, section biography : new orleans saints : saints relocation controversy:
when it became clear that hurricane katrina 's extensive damage to new orleans and the superdome would make it impossible for the saints to play there in 2005, the team temporarily relocated its operations to san antonio and began negotiations to play home games at the alamodome. (the saints, after discussions with the nfl and louisiana state university, eventually agreed to play one "home" game at giants... & article 2001 minnesota vikings season, section preseason : game summaries : week 1: at new orleans saints:
at alamodome, san antonio, texas \\
      \midrule
      \bottomrule
    \end{tabular}
  }
\end{table}

\clearpage
\begin{lstlisting}[caption={Prompt for keyword extraction.},label={lst:keyword_prompt},basicstyle=\ttfamily\footnotesize,breaklines=true]
You are a keyword extraction assistant helping to identify the 
keywords in a question for a co-occurrence search. 
The goal is to check how often the answer to a specific question
(fact) appears in a text corpus. 
To do this, you must identify the keywords in the question that are 
needed to find the fact in the text corpus.

Your job is to analyze a question/answer pair and pull out:
- The minimal term(s) that, when paired with the known answer entities, reliably locate the same fact in a text corpus.
- The goal is to have as few terms as possible while still being able to find the fact.

Guidelines:
- Extract the main keyword from the question that shrinks the search space. 
  E.g., for a song title question, the main keyword is the title of the song.
- Extract additional keywords needed to find the fact in a text corpus. 
  E.g., for a song title, additional keywords are the artist and album. 
- The main keyword should be a single term or short phrase that captures the essence of the question.
- Additional keywords should be a short list of terms (not too long).

Return exactly this JSON (no extra fields or explanation):

{
  "main_keyword":  [string],
  "additional_keywords":  [ string, .. ]
}


Example 1
Input:
Question: "Who were the writers of the song 'Tell Your Heart to Beat Again'?"  
Answer:   "Bernie Herms, Mathew West, Randy Phillips"  
Output:
{
    "main_keyword":  ["Tell Your Heart to Beat Again"],
    "additional_keywords":  ["writers"]
}

Example 2
Input:
Question: "What are the names of recognized dwarf planets in the solar system as of 2024?"  
Answer:   "Ceres, Eris, Pluto, Makemake, Haumea"  
Output:
{
    "main_keyword":  ["dwarf planet"],
    "additional_keywords":  ["solar system"]
}

Example 3
Input:
Question: "What is the legal age of marriage in the United States?"  
Answer:   "18, 19, 21"  
Output:
{
    "main_keyword":  ["marriage"],
    "additional_keywords":  ["legal age", "United States"]
}


Now process the following and produce **only** the JSON:

Question: "{question}"  
Answer:   "{answer}"
\end{lstlisting}

\section{Usage of Large Language Models}

In this work, we used LLMs to polish and rephrase minor sentences of the paper.

\end{document}